\documentclass[twoside]{article}

\usepackage[accepted]{aistats2022}
\usepackage[utf8]{inputenc}
\usepackage[english]{babel}
\usepackage{url}            
\usepackage{hyperref}       
\usepackage{booktabs}       
\usepackage{amsfonts}       
\usepackage{nicefrac}       
\usepackage{microtype}      
\usepackage[dvipsnames, table, svgnames]{xcolor}         

\usepackage{natbib}
\usepackage{amsmath,amssymb,bm}
\usepackage{amsthm}
\usepackage{comment}
\usepackage{cleveref}
\usepackage{amsmath}
\usepackage{tabularx}
\usepackage{multicol, multirow, colortbl}
\usepackage{pifont}
\usepackage{caption}
\usepackage{subcaption}
\usepackage{balance}
\setcitestyle{square}

\usepackage{tikz, pgfplots}

\hypersetup{
    colorlinks=true,
    citecolor = blue,
    linkcolor=blue
}

\newtheorem{theorem}{Theorem}[section]

\newtheorem{lemma}[theorem]{Lemma}
\newtheorem{assumption}[theorem]{Assumption}
\newtheorem{proposition}[theorem]{Proposition}
\theoremstyle{definition}
   \newtheorem{definition}[theorem]{Definition}
   
\theoremstyle{remark}

\DeclareMathOperator*{\argmax}{arg\,max}
\DeclareMathOperator*{\argmin}{arg\,min}

\newcommand{\quotes}[1]{``#1''}

\newcommand\independent{\protect\mathpalette{\protect\independenT}{\perp}}
\def\independenT#1#2{\mathrel{\rlap{$#1#2$}\mkern2mu{#1#2}}}

\usepackage{mathtools}

\DeclarePairedDelimiter\floor{\lfloor}{\rfloor}

\newcolumntype{C}{>{\centering\arraybackslash}X}
\newcommand{\cmark}{\ding{51}}
\newcommand{\xmark}{\textcolor{lightgray}{\ding{55}}}

\newcommand\gauss[2]{1/(#2*sqrt(2*pi))*exp(-((x-#1)^2)/(2*#2^2))}

\begin{document}

\twocolumn[
\aistatstitle{Identifiable Energy-based Representations: An Application to Estimating Heterogeneous Causal Effects}
\aistatsauthor{Yao Zhang$^*$ \And  Jeroen Berrevoets$^*$    \And Mihaela van der Schaar}

\aistatsaddress{University of Cambridge \And University of Cambridge \And University of Cambridge \\ UCLA \\ The Alan Turing Institute}]

\begin{abstract}
Conditional average treatment effects (CATEs) allow us to understand the effect heterogeneity across a large population of individuals. However, typical CATE learners assume all confounding variables are measured in order for the CATE to be identifiable. This requirement can be satisfied by collecting many variables, at the expense of increased sample complexity for estimating CATEs. To combat this, we propose an energy-based model (EBM) that learns a low-dimensional representation of the variables by employing a noise contrastive loss function. With our EBM we introduce a preprocessing step that alleviates the dimensionality curse for {\it any existing} learner developed for estimating CATEs. We prove that our EBM keeps the representations partially identifiable up to some universal constant, as well as having universal approximation capability.
These properties enable the representations to converge and keep the CATE estimates consistent. Experiments demonstrate the convergence of the representations, as well as  show that estimating CATEs on our representations performs better than on the variables or the representations obtained through other dimensionality reduction methods.

\end{abstract}

\vspace{-3mm}
\section{Introduction}\label{sect:intro}

Average treatment effect (ATE) is arguably the most popular estimand in the causal inference literature. With the ATE, one measures if a treatment is effective on average over a population of individuals. However, even if we estimate an ATE accurately, we can not conclude if a treatment is beneficial for a particular individual. In order to get treatment effect estimates for one individual, we condition the ATE on the individual of interest, and arrive at the {\it conditional} average treatment effect (CATE). CATEs know successful applications in areas such as healthcare and education.

While clinical trials represent the gold standard for causal inference, they often have a small number of individuals and narrow inclusion criteria, rendering them unsuitable for use in estimating the causal effects conditional on some  particular individual's confounding variables (covariates). On the other hand, observational datasets are becoming increasingly available, but require careful attention to the biases in the datasets. There is growing interest in leveraging observational data to estimate CATEs, e.g., electronic healthcare records used to determine which patients should get what treatments, or school records to optimize educational policy in low- and high-income communities.

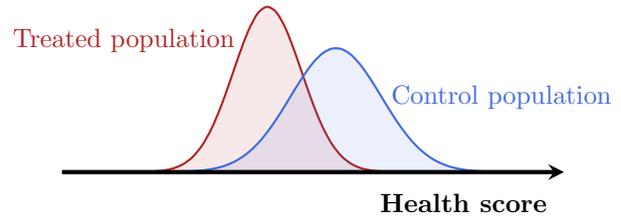
\begin{figure}[t]
    {\centering

    \begin{tikzpicture}
        \begin{axis}[
            every axis plot post/.append style={
            mark=none,domain=-10:10,samples=50,smooth, thick, fill opacity=0.1,},
            xlabel={\bf Health score},
            x label style={at={(axis description cs:0.8,0.15)},anchor=north},
            axis x line=bottom,
            axis y line=none,
            enlargelimits=upper,
            height=4cm,
            width=\columnwidth,
            ticks=none,
            axis line style = ultra thick,
            axis on top,
            clip=false
        ]
        \addplot[color=FireBrick, fill=FireBrick] { \gauss{-1}{1.5}}
        node[pos=0.4, left, fill opacity=1] {Treated population};
        
        \addplot[color=RoyalBlue, fill=RoyalBlue] {\gauss{2}{2}} 
        node[pos=0.7, right, fill opacity=1] {Control population};
            
        \end{axis}

    \end{tikzpicture}
    
    \caption{{\bf Imbalanced treated and control (i.e. untreated) populations.} Individuals with lower health scores are more likely to receive the treatment.}}
    \label{fig:two_popu}
    \vspace{-5pt}
    \rule{.5\linewidth}{.5pt}
    \vspace{-15pt}
\end{figure}

A fundamental assumption for valid causal inference on observational data is called the strong ignorability assumption \citep{rosenbaum1983central,rosenbaum1984reducing}. It assumes independence between the potential outcomes of interest and the treatment variable, conditional on the confounding covariates. Because this assumption is untestable, we often estimate causal effects using all the observed covariates. However, 
estimating CATEs with moderate or high dimensional covariates is challenging. For example, in \Cref{fig:two_popu} we illustrate the treatment assignment process based on one observed covariate, health score. Here, the assignment process creates a discrepancy between the treated and control populations. That is to say, we rarely observe  healthy individuals who receive the treatment, and unhealthy individuals who do not receive the treatment. Then the CATE (i.e., the treatment effect conditional on the health score) becomes difficult to estimate for these individuals. The main reason for this is due to work with finite samples:  the probability of observing two
comparable individuals in a dataset decreases as the covariates dimension increases. However, many covariates in the high-dimensional space are often generated by some common and low-dimensional (latent) variables. Constructing an accurate CATE estimator is then easier on the low-dimensional latents than on high-dimensional observed covariates.

\textbf{Contributions.} In this paper, we explore the assumption that the CATE is a function with an intrinsic dimension lower than the observed covariates.  We propose a representation learning method based on a partially randomized energy-based model (EBM) to embed the covariates into a low-dimensional space before estimating CATEs. This preprocessing step can be used alongside any regression model and learner to reduce their dimensionality curse in CATE estimation. We prove that the representation in the partially randomized EBM is {\it partially identifiable up to some universal constant} for any value of the covariates. Further, the EBM still has {\it universal approximation capability} for estimating any continuous covariates distribution, which avoids excessive information loss from model misspecification. To our best knowledge, identifying representations in deep learning models exactly is still infeasible. Existing theory settles on achieving weaker versions of identifiability with the help of some auxiliary information, e.g., time steps and class labels. Auxiliary information does not exist in most observational datasets. We prove that by optimizing the partially randomized EBM with a noise contrastive loss function and a sample splitting strategy, the representations can converge consistently with increasing sample sizes. 

Experiments on multiple datasets complement our theoretical results. We empirically validate the convergence of the representations with increasing sample sizes. We also show that estimating CATEs based on our representations achieve better performance than directly on the  covariates or the representations obtained via a variety of benchmark dimensionality reduction methods.\footnote{The code of our method is provided at: \url{https://github.com/jeroenbe/ebm-for-cate}. }

\section{Setup}\label{sect:setup}

Following \citep{neyman1923applications} and \citep{rubin1974estimating}, we use the potential outcomes framework to define causal effects. Consider an observational dataset $\mathcal{D} = \left\{O_i = (X_{i},A_{i},Y_{i}):i\in [n]\right\}$, where $[n] = \{1,\dotsc,n\}$. Each individual $i$ is described by a set of covariates $X_{i}\in \mathcal{X}\subseteq \mathbb{R}^{d}$, a binary  treatment variable $A_{i}\in \mathcal{A} = \{0,1\}$ and an observed outcome $Y_{i} \in \mathcal{Y}\subseteq \mathbb{R}$. We assume that the samples in $\mathcal{D}$ are $n$ i.i.d copies of the random variable 
\[
O = (X,A,Y)\sim \mathbb{P}(O) =  \mathbb{P} (Y\mid A, X) \mathbb{P} (A\mid X)  \mathbb{P} (X).
\]

We assume every individual $i$ has two potential outcomes, the control outcome $Y_i(0)$ and the treated outcome $Y_i(1)$. The treatment assignment depends on the individuals’ covariates, i.e., $A_i \not\independent X_i$. This dependence is quantified via the conditional distribution $\pi (X_i) = \mathbb{P}(A_i=1 | X_i)$, also termed as the propensity score in the literature. We make the standard assumptions for causal inference on observational data.

\begin{assumption}[SUTVA, Ignorability and Positivity]
\label{assumption:three}
For any $i\in [N]$ and $a\in \{0,1\}$, $Y_i =Y_i(a)$ if $A_i = a$. For any $i\in [N]$, the distribution of $X_i,A_i,Y_i(0)$ and $Y_i(1)$ satisfies strong ignorability: $Y_i(0),Y_i(1) \independent A_i | X_i$ and positivity:  $\exists~\delta\in (0,1)$ s.t.
  $ \delta < \mathbb{P}(A_i=a | X_i=x) < 1-\delta, \forall x\in \mathcal{X}$ and $a\in \{0,1\}$.
\end{assumption}

We divide $\mathcal{D}$ into a control set and a treated set, $\mathcal{D}_{\text{c}} = \{( X_i,A_i,Y_i ): A_i=0,i\in[n]\}$ and $\mathcal{D}_{\text{t}} = \{( X_i,A_i,Y_i ):A_i=1,i\in[n]\}$.  We denote the sample sizes of $\mathcal{D}_{\text{c}}$ and $\mathcal{D}_{\text{t}}$ by $n_{\text{c}}=|\mathcal{D}_{\text{c}}|$ and $n_{\text{t}}=|\mathcal{D}_{\text{t}}|$. Under \Cref{assumption:three},  $\mu_{a}(x) := \mathbb E\{Y(a)\mid X=x\}=\mathbb E\{Y\mid X=x, A=a\} $ for $a\in \{0,1\}$. Then we can  identify the conditional average treatment effect (CATE) $ \tau(x)$ by
\begin{equation}\label{equ:T_identification}
    \tau(x) = \mathbb E [Y (1) - Y (0) | X = x] = \mu_{1}(x) - \mu_{0}(x).
\end{equation}
In nonparametric regression, the dimension and smoothness of the data generating function determine the expected squared error of a regression model \citep{stone1980optimal}. The error of the used regression model determines the error of a CATE learner.

\begin{definition}[H\"older ball]
The H\"older ball $\mathcal{H}_d(s)$ is the set of functions $f: \mathbb{R}^d \rightarrow \mathbb{R}$ supported on $\mathcal{X}\subseteq \mathbb{R}^d $ with their partial derivatives satisfying that
\[\left| \frac{\partial^m f}{\partial^{m_1}\cdots \partial^{m_d}} (x)-\frac{\partial^m f}{\partial^{m_1}\cdots \partial^{m_d}}(x')\right|\lesssim \|x-x'\|_2^{s-\floor{s}},\]
$\forall x,x'\in \mathcal{X}$ and  $m = (m_1,\cdots,m_d)$ s.t. $\sum_{j=1}^{d}m_j=\floor{s}$.
\end{definition}
The notation $a\lesssim b$ denotes the relation $a\leq  C b$ for some universal constant $C$. Essentially, $\mathcal{H}_d(s)$ is the class of smooth  functions that are close to their $\floor{s}$-order Taylor approximations. We assume $\mu_0, \mu_1$, $\pi$ and $\tau$ are $s$-smooth functions in the H\"older balls $\mathcal{H}_d(s)$ for some non-negative smoothness parameter $s=\alpha_0,\alpha_1,\beta, \gamma$, respectively. 


The identification formula \eqref{equ:T_identification} motivates a common estimation strategy called a \quotes{T-learner}, where \quotes{T} refers to \quotes{Two} regression models. A T-learner estimates $\mu_{0}$ and $\mu_{1}$ by fitting two separate regression models, $\hat{\mu}_{0}$ and $\hat{\mu}_{1}$, on $\mathcal{D}_{\text{c}}$ and $\mathcal{D}_{\text{t}}$, respectively. It estimates the CATE as the difference $\hat{\tau}(\cdot) = \hat{\mu}_{1}(\cdot) - \hat{\mu}_{0}(\cdot)$. 
Suppose the expected squared error of $\hat \mu_0$ and $\hat \mu_1$ are
$n_{\text{c}}^{-\frac{2\alpha_0}{2\alpha_0+d}}$ and $n_{\text{t}}^{-\frac{2\alpha_1}{2\alpha_1+d}}$, respectively.
A T-learner's expected squared error $\mathbb{E}\left\{[\hat{\tau}(X)- \tau(X)]^2\right\}$
is 
$O( n_\text{c}^{-\frac{2\alpha_0}{2\alpha_0+d}}+  n_{\text{t}}^{-\frac{2\alpha_1}{2\alpha_1+d}}).$

There are other advanced learners based on different identification formulas, e.g., X-learner \citep{kunzel2019metalearners}, R-learner \citep{nie2021quasi} and DR-learner \citep{kennedy2020optimal}. For example, the identification formula of a DR-learner is based on the uncentered first-order influence function $\phi(x)$ of the ATE,
\begin{equation}\label{equ:if}
\begin{split}
\phi(x) & = \frac{A}{\pi(x)}\left[ Y - \mu_1(x)\right]+ \mu_1(x) \\
&\hspace{1.3cm} - \frac{1-A}{1-\pi(x)}\left[Y - \mu_0(x)\right]- \mu_0(x).
\end{split}
\end{equation}
Splitting $\mathcal{D}$ into three subsets $\mathcal{D}_1$, $\mathcal{D}_2$ and $\mathcal{D}_3$, a DR-learner estimates the CATE as follows: estimate $
\mu_{0}$ and $\mu_{1}$ on $\mathcal{D}_1$, estimate $\pi$ on $\mathcal{D}_2$, then estimate the CATE $\tau(x)$ by regressing $\hat{\phi}(X)$ onto $X$ in $\mathcal{D}_3$, where $\hat{\phi}(X)$ is generated by plugging the estimators $\hat{\mu}_{0}, \hat{\mu}_{1}$ and $\hat{\pi}$ into the expression \eqref{equ:if}.
\citet[Theorem 2]{kennedy2020optimal} shows that
a DR-learner's expected squared error is
\[
  O(
\tilde{n}_2^{-\frac{2\beta}{2\beta +d}} \big(
\tilde{n}_{1,\text{c}}^{-\frac{2\alpha_0}{2\alpha_0+d}} +  \tilde{n}_{1,\text{t}}^{-\frac{2\alpha_1}{2\alpha_1+d}}
\big) + \tilde{n}_3^{-\frac{2\gamma}{2\gamma+d}}
),  
\]
where $\tilde{n}_m = |\mathcal{D}_m|, m=1,2,3$, $\tilde{n}_{1,\text{c}}$ and $\tilde{n}_{1,\text{t}}$ are the number of control and treated individuals in $\mathcal{D}_1$. The efficiency loss from sample splitting can be remedied by cross-fitting \citep{nie2021quasi,chernozhukov2018double}. DR-learner can improve CATE estimation by leveraging the smoothness of $\tau$. But for an accurate CATE estimator to exist in finite samples, the requirement on the smoothness parameters ($\alpha_0,\alpha_1,\beta$ and $\gamma$) is restrictive if the amount of dimensions $d$ is large, regardless of which learner we use. In this paper, we focus on improving CATE learners by reducing $d$, which is plausible under the following assumption.
\begin{assumption}\label{assumption:dim}
For some positive integer $d^* < d$, there exists some variable $U \in \mathcal{U}\subseteq \mathbb{R}^{d^*}$ that generates the variables $X,A$ and $Y$, i.e., the data distribution $\mathbb{P}(O)$  has a density function $p_O(o)$ which satisfies that for any $o\in \mathcal{O}=\mathcal{X}\times \mathcal{A}\times \mathcal{Y}$, $p_O(o)$ is given by
\[
\int_{\mathcal{U}}p_{Y\mid A,U}(y\mid a, u)  p_{A\mid U}(a\mid  u) p_{X\mid U}(x\mid  u )  p_{U}( u) du.
\]
\end{assumption}
This assumption is realistic as  observational data often includes covariates that represent the same aspect of an individual. For example, an individual's health status can be represented by some collection of covariates, e.g., some disease-specific symptoms. These covariates are correlated and contain overlapping information about the individual. Under \Cref{assumption:dim}, both the outcome $Y$ and treatment $A$ are generated by these abstract aspects $U$. This hints at a potentially more sample-efficient estimation strategy: first learning these aspects as a low-dimensional representation of the covariates, then fitting $\hat{\mu}_0,\hat{\mu}_1$ (and $\hat{\pi}$) on the low dimensional representation to estimate the CATE.

\textbf{Why CATE as an application?}
The limitation of representation learning is that the representation itself is non-smooth and takes many samples to learn. In supervised learning on a fully labelled dataset, there is no obvious advantage of learning the representations first over learning the label directly. By contrast, observational datasets for CATE estimation are often imbalanced ($n_t  \ll n_c$) so that the term $n_{\text{t}}^{-\frac{2\alpha_1}{2\alpha_1+d}}$ in a T-learner's expected squared error is very large. Leveraging the smoothness of $\tau$ is not  very effective in the presence of high-dimensional covariates. Rather than directly using all the covariates to construct $\hat{\mu}_0,\hat{\mu}_1$ (and $\hat{\pi}$), we employ representation learning. In particular, because the representation can be learnt using {\it all the samples}, i.e., $n_t+n_c$ samples from both the treated and control group. Then by using the low dimensional representation to estimate CATEs, the learners  will potentially have smaller expected errors.

In the next section, we present a deep neural network as our representation learning model, which allows us to keep the nonparametric merit of the regression models built on top of the representation. Under \Cref{assumption:three}, we consider the representation learning model, concatenated together with the outcome (propensity score) model, as an outcome (propensity score) model {\it based on the observed covariates}. The consistency of the resulting CATE estimator thus depends on the consistency of {\it both} models. This requires the representation to be identifiable, which is so far still impossible to achieve exactly for overparameterized neural networks. In our next section, we provide an approximate solution to this problem, sufficient for CATE estimation.

\section{Model}\label{sect:ebm}

A parameter is identifiable in a class of statistical models if every model describing the same distribution, has the same value of the parameter. If models with different parameter-values give the same distribution, i.e., generate the same observed data in the large data limit, we can no longer find the true model from the data even if the sample size is large \citep{lewbel2019identification}.


Generally, identifiability is often achieved by introducing some constraint on the model class, as is also the case here. We will construct partially identifiable representations in a class of partially randomized energy-based models (EBMs). By partially identifiable, we mean if two models give the same distribution, then their representations are only different by some universal constant. A partially randomized EBM is constructed in the two steps detailed below.



%

%

\textbf{Step 1.} Suppose we want to learn a $k$-dimensional representation of the covariates $(k<d)$\footnote{The errors of CATE learners depend on the performance of the outcome and propensity score models $\hat{\mu}_0$, $\hat{\mu}_1$ and $\hat{\pi}$. Because some CATE learners do not use a propensity score model, the dimension $k$ is tuned as a hyper-parameter via cross-validation on the observed outcomes in this paper.}. We let
$f_{\theta}:\mathcal{X}\rightarrow \mathbb R^{k}$ be a neural network that generates the $k$-dimensional data representation.
We define $k$ standard EBMs \citep{lecun2006tutorial} on $\mathcal{X}$ with a {\it shared} representation $f_{\theta}$:
\begin{equation}\label{equ:model}
   p_{\theta,j}(x) =Z_{\theta,j}^{-1}\exp\left[-\beta_j^{\top}  f_{\theta}(x)\right],~\forall j\in [k], 
\end{equation}
where 
$Z_{\theta,j} =\int_{\mathcal{X}} \exp\left[-\beta_j^{\top}  f_{\theta}(x)\right] dx$. The number of standard EBMs is the same as the size of the representation for a purpose. Roughly speaking,
we want to create $k$ equations to determine a $k$-dimensional representation (for details, we refer to \Cref{prop:id} and its proof in \Cref{sect:app:proof33}). 
The EBM \eqref{equ:model} can be written as an exponential family distribution,
$p_{\theta,j}(x)= h(x)\exp \left[\lambda_{j}^{\top }f_{\theta}(x)  - \psi_{\theta,j}          \right],$
where $h(x)=1$ and $\lambda_j = -\beta_j$, and $ \psi_{\theta,j}=\log\left(\int_{\mathcal{X}}h(x) \exp    \left[\lambda_{j}^{\top }f_{\theta}(x)\right] dx \right)$.

\begin{proposition}\label{prop:suff}
For every $j\in [k]$,
 $f_{\theta}$ is a minimum and sufficient statistic in model \eqref{equ:model}.
\end{proposition}

\textbf{Step 2.} Let $\mathcal{P}= \{p_{\theta,j}\mid  \beta_j\in  \mathbb{R}^{k}, \theta\in \Theta \}$ denote the space of standard EBMs, and $\mathcal{P}(\beta_j)$ denote the  subset of $\mathcal{P}$ with $\beta_j$ fixed. Let $B = \left(\beta_1,\dotsc,\beta_k\right)$ be the  $k\times k$ matrix whose $j$-th column is $\beta_j$. A partially randomized EBM is given by multiple standard EBMs with a shared representation $f_{\theta}$ and a fixed random orthogonal matrix $B$, i.e., $\theta$ is the only learnable parameter.

\begin{definition}[Partially Randomized EBM]
A partially randomized EBM is given by 
\begin{equation}\label{equ:prebm}
  p_{\theta} = (p_{\theta,j}:j\in [k])\in \bigtimes_{j=1}^{k}\mathcal{P}(\beta_j),
\end{equation}
where $B = \left(\beta_1,\dotsc,\beta_k\right)$ is a $k\times k$ random orthogonal matrix s.t. $B B^{\top } = I_{k\times k} $.
\end{definition}

One easy approach to construct $B$ is to first generate a random matrix $B_0\in \mathbb{R}^{k\times k}$, where each entry is drawn independently from a standard normal distribution, and then taking $B$ as the matrix of eigenvectors of $B_0$. The partially randomized EBM satisfies the  partial identifiability defined as follows.

\begin{proposition}\label{prop:id}
For any $k\times k$ random orthogonal matrix $B=(\beta_1,\dotsc,\beta_k)$ and  $p_{\theta,j}, p_{\tilde{\theta},j} \in \mathcal{P}(\beta_j)$ such that
$p_{\theta,j}(\cdot) =  p_{\tilde{\theta},j}(\cdot),\forall j\in [k]$, we have
\begin{equation}\label{equ:id}
 f_{\theta}(\cdot)  - f_{\tilde{\theta}}(\cdot) = C\ \text{ for some constant vector $C$. }
\end{equation}
\end{proposition}
Perhaps surprisingly, the randomization strategy above does not overly decrease the model complexity. \Cref{prop:approx} verifies the universal approximation capability of the partially randomized EBM. The proof is attained by showing that $\mathcal{P}(\beta_j)$ satisfies the conditions in the Stone-Weierstrass approximation theorem. The proofs of all the propositions can be found in \Cref{sect:proof}.

\begin{proposition}\label{prop:approx}
For any continuous density function $p_X:\mathcal{X}\rightarrow \mathbb{R}^+$, $k\times k$ random orthogonal matrix $B$, and $\epsilon>0$, there exists $p_{\theta,j} \in \mathcal{P}(\beta_j)$ such that  $\sup_{x\in \mathcal{X}} |p_X(x) -  p_{\theta,j}(x)|\leq \epsilon$ for all $j\in [k]$. 
\end{proposition}
Next, we will introduce a training strategy for our partially randomized EBM, which will enable the learnt representation model to converge to a limiting set in which any functions are only different by some constants $C$ like 
$f_{\theta}$ and $f_{\tilde{\theta}}$ in \eqref{equ:id}.
This will enable the follow-up CATE estimates to converge consistently because the regression models $\hat{\mu}_0,\hat{\mu}_1$ (and $\hat{\pi}$) are indifferent to conditioning on a random variable, or the same random variable plus some constant vector. Furthermore, by standardizing the learnt representations, we can fix their mean to 0 and their variance to 1 in any sample size. Given that the representations have mean 0 and variance 1, by partial identifiability and consistency, the representations obtained from different runs of the experiments will have a correlation close to 1 at each dimension in large samples, as will be demonstrated experimentally in \Cref{sect:6.2}.

\section{Method}\label{sect:method}

Fitting energy-based models (EBMs) by maximum likelihood estimation (MLE) is often infeasible because the partition function ($Z_{\theta, j}$) is intractable.  Noise Contrastive Estimation (NCE) proposed by \citep{gutmann2010noise,gutmann2012noise} is a consistent and computationally efficient alternative. The high-level idea of NCE is to optimize an EBM by contrasting it with another noise distribution with known and easy-to-sample density. Advanced methods have been proposed to tune the noise distribution, see for example \citet{gao2020flow,bose2018adversarial,Ceylan2018ConditionalNE}.

Here for every individual $i\in [n]$, we draw $b$ corrupted samples $\tilde{X}_{i1},\dotsc, \tilde{X}_{ib}$ from a noise distribution $p_{\tilde X\mid X}(\tilde{x}\mid X_i)$ defined as follows. Each $\tilde{X}_{ia}$ is generated in two steps: (1) we sample an independent 
binary variable with some probability for each feature of $X_i$, used to decide which features of $X_i$ will be corrupted, then (2) corrupt each  selected continuous feature by adding white noise drawn from a standard normal distribution, and corrupt each selected categorical feature by uniformly sampling a value from its range. A mathematical description of $p_{\tilde X\mid X}(\tilde{x} \mid X_i)$ is provided in  \Cref{sect:noise}. Overall, the original and corrupted data of individual $i$ is given by
\[
\bar{X}_i = (X_i,\tilde{X}_{i1},\dotsc, \tilde{X}_{ib} )\sim   p_{X}(x)\prod_{a=1}^{b}p_{\tilde X\mid X}(\tilde{x} \mid x).
\]

We split the $n$ individuals into $k$ subsets
$\mathcal{I}_j, j\in [k]$, to train each of of $k$ models $p_{\theta,j}(x)$ in the partially randomized EBM as in \eqref{equ:prebm}. Suppose we randomly permute the columns of $\bar{X}_i$ and let $V_i = (V_{ia} :a\in [b+1])$ be the permuted $\bar{X}_i$. Then each column of $V_i$ has equal probability $(b+1)^{-1}$ for being the original sample $X_i$. We derive the predictive probability of $V_{ia} = X_i$ from the posterior distribution,
\begin{equation}\label{equ:poster_prob}
\begin{split}
q_{\theta,j}(a \mid V_i)  = &  \frac{(b+1)^{-1} p_{\theta,j}(V_{ia})\tilde{p}_{-a}(V_i)}{\sum_{c=1}^{b+1}(b+1)^{-1}p_{\theta,j}(V_{ic})\tilde{p}_{-c}(V_i)},  \\
 \end{split}
\end{equation}
where $\tilde{p}_{-a}(V_i) =\prod_{a\in [b+1]:a'\neq a}\tilde{p}_{\tilde X\mid X}(V_{ia'}\mid V_{ia})$. It is noteworthy that the intractable partition function $Z_{\theta, j}$ in $p_{\theta,j}$ (in \eqref{equ:model}) cancels out in the expression of  $q_{\theta,j}(a \mid V_i)$.
Let $W_i\in \{0,1\}^{b+1}$ indicate which column of $V_{i}$ is $X_i$. We can think of $\{(V_i,W_i): i:\in \mathcal{I}_j\} $ as a set of labeled \quotes{images} and optimize the probability $q_{\theta,j}(a \mid V_i)$ to predict $W_i$. Let $n_j=|\mathcal{I}_j|$. Our objective function is the negative cross-entropy\footnote{This is essentially  the ranking objective in \citep{Jzefowicz2016ExploringTL,Ma2018NoiseCE} with a different noise distribution. We reformulate the training strategy as a more intuitive multiclass classification task.},
$ \mathcal{L}_{n}(\theta)  = k^{-1} \sum_{j=1}^{k} \mathcal{L}_{n,j}(\theta)$, where $\mathcal{L}_{n,j}(\theta)   $ is given by 
\begin{equation}\label{equ:rank}
\begin{split}
\mathcal{L}_{n,j}(\theta) = &   n_j^{-1}\sum_{i\in \mathcal{I}_j} \sum_{a=1}^{b+1} W_{ia}\log q_{\theta,j}(a \mid V_i)       \\
= & n_j^{-1}\sum_{i\in \mathcal{I}_j} \log q_{\theta,j}(1 \mid \bar{X}_i).
\end{split}
\end{equation}
 The representation model $f_{\theta}$ is trained on all the samples, even though we split the samples across the models $p_{\theta,j},j\in[k],$ in our partially randomized EBM.

The training strategy here follows the same principle as the other representation learning methods, e.g., \citep{Vincent2010StackedDA,vincent2011connection}: 
assume the covariates $X_i$ live in some $d^*$-dimensional manifold $(d^*<d)$. If $q_{\theta,j}(a \mid V_i)$  is predictive of $W_i$, i.e., can distinguish any true sample $X_i\sim p_{X}(x)$ from its noisy proxies $\tilde{X}_{ia}\sim \tilde{p}_{\tilde X\mid X}(\tilde{x}\mid X_i)$, we have $p_{\theta,j}(x)\approx p_X(x)$ in \eqref{equ:poster_prob}. This implies that the low-dimensional representation given by $f_{\theta}(x)$  is 
informative of the true covariates $X_i$; the representation is also predictive of the outcome and treatment because they are generated by the covariates.

\Cref{prop:consistent} below shows that by our training strategy, the learnt parameter $\hat{\theta}_{n}$ will converge to a set of limits $\Theta_0$ s.t. $ p_{\theta_0,j}(x) = p_X(x)$ for any $x\in \mathcal{X}$ and $\theta_0\in \Theta_0$.
Then by \eqref{equ:id} in \Cref{prop:id}, no matter which $\theta_0\in \Theta_0$ that  $\hat{\theta}_n$  converges to,  the limit of $f_{\hat{\theta}_n}(x)$ will be only different by some universal constant. 

\begin{proposition}\label{prop:consistent}
Suppose that the covariates space $\mathcal{X}$ is a compact subset of $ \mathbb{R}^{d}$, $f_{\theta}(x)$ has a compact parameter space $\Theta$, and $f_{\theta}(x)$ is continuous with respect to its parameter $\theta$ for any $x\in \mathcal{X}$. For any $k\times k$ random orthogonal matrix $B$,
under \Cref{prop:approx}, we assume for any continuous density function $p_{X}(x)$ defined on $\mathcal{X}$, there exists a countable subset $\Theta_0\subset \Theta$ s.t.
$ p_{\theta_0,j}(x) = p_{X}(x)$ for any $x\in \mathcal{X}$ and $\theta_0\in \Theta_0$.  For any number of noise samples $b$ and
$\hat{\theta}_n \in \argmax_{\theta\in\Theta} \mathcal{L}_{n}(\theta) $, we have $\lim_{n\rightarrow\infty}\hat{\theta}_n \in   \Theta_0 $ with probability 1. 
\end{proposition}

Essentially, both MLE and NCE are special cases of M-estimators in statistics \citep{van2000asymptotic}. The proposition is proven by showing 
that $ \mathcal{L}_{\infty,j}(\theta)$ is maximized by $q_{\theta,j}(a \mid V_i)$ with $p_{\theta,j}(x)= p_X(x)$, and the standard conditions for consistent M-estimators hold for $\hat{\theta}_n $  under a weaker identifiability assumption; see \Cref{sect:consistent} for more details. 

\section{Related works}\label{sect:related}
Here we provide related works on three different areas.

\textbf{Identifiability theory.} \citet{khemakhem2020ice} propose two definitions of identifiability for EBMs; weak and strong identifiability (in their Definitions 1 and 2). Their EBM is more complex than ours with $\beta_j$ as a learnable parameter, while their objective is to identify both $\beta_j$ and $f_{\theta}(x)$. This is unnecessary for the application in our paper. Arguably, the partial identifiability defined in our paper is stronger than both of their definitions. In their strong identifiability, under some assumptions, each dimension of $f_{\theta}$ is identifiable up to be multiplied by and plus some constants, and each dimension of $f_{\theta}$ can be permuted in any order. They also require a specific network architecture for $f_{\theta}$. In our work, we use a {\it simpler} partially randomized EBM to achieve a {\it stricter} version of identifiability, without sacrificing the approximation capability of the EBM or restricting the architecture of  $f_{\theta}$.

The works on nonlinear ICA and its generalization \citep{hyvarinen2016unsupervised,hyvarinen2019nonlinear,khemakhem2020variational,mita2021identifiable} propose the idea of using contrastive learning for identifiable feature extraction when some auxiliary information (e.g., time steps) about the features is available. We use sample splitting and a noise contrastive loss function for training the partially randomized EBM, assuming {\it no auxiliary information} is provided in the observational data. 
\citet{monti2020causal} and  \citet{wu2020causal} propose non-linear ICA based methods for causal inference 
on structural causal models \citep{pearl2009causality}. The setup and problems studied in their papers are different from our method which is developed within the potential outcomes framework. 

\textbf{Representation learning.} Representation learning is recently applied to balance or match the covariate distribution between the treated and control group in observational data, by minimizing the distributional distance between the group \citep{shalit2017estimating}, preserving local similarity \citep{yao2018representation}, minimizing counterfactual variance \citep{zhang2020learning} and adversarial training \citep{kallus2020deepmatch}. We note that supervised dimensionality reduction in a deep learning model is not reliable because the model can easily overfit the limited outcome data without finding an informative representation of the covariates.  Our proposed method works more generally as a preprocessing step to reduce the dimentionality curse for any regression model, including these deep learning models which balance the distribution in one of their hidden layers.

In statistics, sufficient dimensionality reduction (SDR) \citep{lee2013general,cook2009regression,li1991sliced,adragni2009sufficient}
has been used in the models for estimating ATE and CATE \citep{Huang2020RobustIO,luo2019learning,cheng2020sufficient,ma2019robust,ghosh2018sufficient}. If the subspace spanned by the columns of a $d\times k$ matrix $\theta$ with $k\leq d$ satisfies that $Y\independent X\mid \theta^{\top}X$, we call this subspace a SDR subspace. The idea of SDR is to project the covariates $X$ onto this subspace before feeding it into a parametric or nonparametric regression model to estimate $Y$. To achieve the desired conditional independence, $\theta$ is jointly learnt with the regression model. 
This is not straightforward for some of the ML models. e.g., decision tree.
\citet{kallus2018causal} proposes a matrix factorization based method for preprocessing noisy and missing covariates.
In contrast with these methods, our method performs nonlinear dimensionality reduction of the covariates, which is more general for the data living in some low-dimensional manifold, including linear subspace. \citet{Nabi2017SemiParametricCS} and \citet{Berrevoets2020OrganITEOT} propose methods to deal with high-dimensional treatment variables, which is not the problem considered in our paper. 

\textbf{Covariates selection.}  When there are irrelevant covariates in a dataset, data analysis should start with a covariates selection method, e.g., \citep{de2011covariate,shortreed2017outcome,greenewald2021high}. However, covariates selection methods often have no guarantee to find the correct adjustment set for causal inference in finite samples. Furthermore, (selected) covariates are correlated, especially when we allow more covariates to be selected in order to satisfy the strong ignorability assumption. On the basis of this, our representation learning method can be applied to further reduce the dimensionality of the correlated covariates and improve the accuracy of CATE estimation. In general, covariates selection and our method are applied in different stages and complement each other in the data analysis process.

\section{Experiments}\label{sect:experiments}
We make two claims in our paper: (1) using our method as a preprocessing step increases the performance of CATE learners; (2) the representation in our model \eqref{equ:prebm} is partially identifiable so that the
learnt representations and downstream CATE estimates are consistent. We test these two claims in the following subsections. Throughout our experiments, we use four different CATE learners: X-Learner, DR-learner, T-Learner, and R-learner \citep{econml}. We provide more details of our experiments (e.g., on learners and hyperparameters) in \Cref{sect:hyper}.

\renewcommand{\arraystretch}{1.0}
\setlength{\tabcolsep}{1.5pt}
\begin{table*}[t]
    \caption{{\bf Results on synthetic data and semi-synthetic data (Twins).} Each row reports the average PEHE (lower is better) over ten runs for each CATE learner (standard deviation in scriptsize): both {\it with} representations (indicated as \quotes{\cmark}), and {\it without} representation (indicated as \quotes{\xmark}). For each run, we learn a new representation. In the above two blocks, we vary sample sizes and dimensions using our synthetic setup, and in the bottom block we vary the sample size for the Twins-dataset.  Using our EBM yields superior testing performance for a range of CATE learners (indicated in bold). In \colorbox{green!10}{green}, we emphasize the best results per row, each time {\it with} EBM. While there may be duplicate values, we highlight only those that are best beyond the rounding applied here.}
    \vspace{-13pt}
    \label{tab:res:CATE:synth}
\begin{center}
    \begin{tabularx}{\textwidth}{l r @{\hskip 2pt}  *{4}{|CC}}
    
    \toprule
   \multicolumn{2}{r|}{Methods}   & \multicolumn{2}{c}{\bf X-Learner}  
      & \multicolumn{2}{c}{\bf DR-Learner} 
      & \multicolumn{2}{c}{\bf T-Learner} 
      & \multicolumn{2}{c}{\bf R-Learner}\\
  \midrule     
 \multicolumn{2}{r|}{EBM} & \xmark & \cmark & \xmark & \cmark & \xmark & \cmark & \xmark & \cmark \\

    \toprule

      $d$ & $n$& 
    \multicolumn{8}{c}{\it Synth. data with increasing sample size and increasing dimensions}\\
     \midrule   
    
    $50$ &$100$ & 
    \footnotesize{2.309~\scriptsize{$\pm$.00}} & {\footnotesize{\bf 1.994}~\scriptsize{$\pm$.02}} & 
    \footnotesize{4.594~\scriptsize{$\pm$.56}} & {\footnotesize{\bf 2.017}~\scriptsize{$\pm$.04}} & 
    \footnotesize{2.441~\scriptsize{$\pm$.00}} & {\footnotesize{\bf 1.993}~\scriptsize{$\pm$.01}} & 
    \footnotesize{3.194~\scriptsize{$\pm$.26}} & {\cellcolor{green!10} \footnotesize{\bf 1.982}~\scriptsize{$\pm$.04}} \\
    
    $100$ &$250$ & 
    \footnotesize{2.779~\scriptsize{$\pm$.00}} & {\footnotesize{\bf 2.018}~\scriptsize{$\pm$.01}} &  
    \footnotesize{4.056~\scriptsize{$\pm$.32}} & {\footnotesize{\bf 2.154}~\scriptsize{$\pm$.39}} & 
    \footnotesize{2.838~\scriptsize{$\pm$.00}} & {\footnotesize{\bf 2.019}~\scriptsize{$\pm$.01}} & 
    \footnotesize{3.702~\scriptsize{$\pm$.23}} & {\cellcolor{green!10}\footnotesize{\bf 2.018}~\scriptsize{$\pm$.01}} \\
    
     $150$ &$500$ &
    \footnotesize{2.618~\scriptsize{$\pm$.00}} & {\footnotesize{\bf 2.000}~\scriptsize{$\pm$.01}} & 
    \footnotesize{3.030~\scriptsize{$\pm$.12}} & {\footnotesize{\bf 2.001}~\scriptsize{$\pm$.01}} & 
    \footnotesize{2.641~\scriptsize{$\pm$.00}} & {\cellcolor{green!10}\footnotesize{\bf 2.000}~\scriptsize{$\pm$.01}} & 
    \footnotesize{2.877~\scriptsize{$\pm$.08}} & {\footnotesize{\bf 2.000}~\scriptsize{$\pm$.01}} \\
    
    $200$ &    $1$k & 
    \footnotesize{2.185~\scriptsize{$\pm$.00}} & {\footnotesize{\bf 1.940}~\scriptsize{$\pm$.01}} & 
    \footnotesize{2.283~\scriptsize{$\pm$.02}} & {\footnotesize{\bf 1.941}~\scriptsize{$\pm$.01}} & 
    \footnotesize{2.189~\scriptsize{$\pm$.00}} & {\cellcolor{green!10}\footnotesize{\bf 1.939}~\scriptsize{$\pm$.01}} & 
    \footnotesize{2.271~\scriptsize{$\pm$.01}} & {\footnotesize{\bf 1.940}~\scriptsize{$\pm$.01}} \\
    
     $250$ &$1.5$k &
    \footnotesize{2.267~\scriptsize{$\pm$.00}} & {\footnotesize{\bf 1.949}~\scriptsize{$\pm$.02}} & 
    \footnotesize{2.427~\scriptsize{$\pm$.01}} & {\footnotesize{\bf 1.976}~\scriptsize{$\pm$.00}} & 
    \footnotesize{2.271~\scriptsize{$\pm$.00}} & {\cellcolor{green!10}\footnotesize{\bf 1.948}~\scriptsize{$\pm$.01}} & 
    \footnotesize{2.436~\scriptsize{$\pm$.02}} & {\footnotesize{\bf 1.949}~\scriptsize{$\pm$.02}} \\
    
    \midrule
    &$n$& 
    \multicolumn{8}{c}{\it Synth. data with increasing sample size and dimensions fixed at $d=100$}\\
    \midrule
    
     &$100$& 
    \footnotesize{2.134~\scriptsize{$\pm$.00}} & {\footnotesize{\bf 1.927}~\scriptsize{$\pm$.01}} & 
    \footnotesize{24.61~\scriptsize{$\pm$9.9}} & {\footnotesize{\bf 2.096}~\scriptsize{$\pm$.09}} & 
    \footnotesize{2.279~\scriptsize{$\pm$.00}} & {\footnotesize{\bf 1.929}~\scriptsize{$\pm$.01}} & 
    \footnotesize{3.192~\scriptsize{$\pm$.13}} & {\cellcolor{green!10}\footnotesize{\bf 1.925}~\scriptsize{$\pm$.01}} \\
    
     &$250$&
    \footnotesize{2.779~\scriptsize{$\pm$.00}} & {\footnotesize{\bf 2.018}~\scriptsize{$\pm$.01}} &  
    \footnotesize{4.056~\scriptsize{$\pm$.32}} & {\footnotesize{\bf 2.154}~\scriptsize{$\pm$.39}} & 
    \footnotesize{2.838~\scriptsize{$\pm$.00}} & {\footnotesize{\bf 2.019}~\scriptsize{$\pm$.01}} & 
    \footnotesize{3.702~\scriptsize{$\pm$.23}} & {\cellcolor{green!10}\footnotesize{\bf 2.018}~\scriptsize{$\pm$.01}} \\
     
    &$500$ &
    \footnotesize{2.155~\scriptsize{$\pm$.00}} & {\footnotesize{\bf 2.056}~\scriptsize{$\pm$.02}} & 
    \footnotesize{2.334~\scriptsize{$\pm$.07}} & {\footnotesize{\bf 2.273}~\scriptsize{$\pm$.67}} & 
    \footnotesize{2.166~\scriptsize{$\pm$.00}} & {\cellcolor{green!10}\footnotesize{\bf 2.053}~\scriptsize{$\pm$.02}} & 
    \footnotesize{2.271~\scriptsize{$\pm$.05}} & {\footnotesize{\bf 2.056}~\scriptsize{$\pm$.02}} \\

    &$1$k  &
    \footnotesize{2.059~\scriptsize{$\pm$.00}} & {\cellcolor{green!10}\footnotesize{\bf 1.964}~\scriptsize{$\pm$.02}} &
    \footnotesize{2.105~\scriptsize{$\pm$.01}} & {\footnotesize{\bf 2.016}~\scriptsize{$\pm$.16}} & 
    \footnotesize{2.061~\scriptsize{$\pm$.00}} & {\footnotesize{\bf 1.964}~\scriptsize{$\pm$.02}} & 
    \footnotesize{2.086~\scriptsize{$\pm$.01}} & {\footnotesize{\bf 1.965}~\scriptsize{$\pm$.02}} \\

    &$1.5$k &
    \footnotesize{2.013~\scriptsize{$\pm$.00}} & {\footnotesize{\bf 1.998}~\scriptsize{$\pm$.02}} & 
    \footnotesize{2.043~\scriptsize{$\pm$.01}} & {\footnotesize{\bf 1.998}~\scriptsize{$\pm$.02}} & 
    \footnotesize{2.014~\scriptsize{$\pm$.00}} & {\footnotesize{\bf 1.998}~\scriptsize{$\pm$.02}} & 
    \footnotesize{2.024~\scriptsize{$\pm$.01}} & {\cellcolor{green!10}\footnotesize{\bf 1.991}~\scriptsize{$\pm$.02}} \\

    \midrule
    &$n$& 
    \multicolumn{8}{c}{\it Twins ($d=48$)  with increasing sample size}\\
    \midrule
    
    &$500$ &             
    \footnotesize{0.214~\scriptsize{$\pm$.00}} & {\cellcolor{green!10}\footnotesize{\bf 0.144}~\scriptsize{$\pm$.00}} & 
    \footnotesize{0.236~\scriptsize{$\pm$.04}} & {\footnotesize{\bf 0.182}~\scriptsize{$\pm$.05}} & 
    \footnotesize{0.221~\scriptsize{$\pm$.00}} & {\footnotesize{\bf 0.145}~\scriptsize{$\pm$.00}} & 
    \footnotesize{0.222~\scriptsize{$\pm$.02}} & {\footnotesize{\bf 0.145}~\scriptsize{$\pm$.00}} \\
    
    &$1$k &
    \footnotesize{0.294~\scriptsize{$\pm$.00}} & {\footnotesize{\bf 0.162}~\scriptsize{$\pm$.00}} & 
    \footnotesize{0.348~\scriptsize{$\pm$.12}} & {\footnotesize{\bf 0.173}~\scriptsize{$\pm$.03}} & 
    \footnotesize{0.301~\scriptsize{$\pm$.00}} & {\footnotesize{\bf 0.162}~\scriptsize{$\pm$.01}} & 
    \footnotesize{0.532~\scriptsize{$\pm$.11}} & {\footnotesize{\cellcolor{green!10}\bf 0.161}~\scriptsize{$\pm$.00}} \\
    
    &$1.5$k&
    \footnotesize{0.165~\scriptsize{$\pm$.00}} & {\footnotesize{\bf 0.154}~\scriptsize{$\pm$.00}} & 
    \footnotesize{0.189~\scriptsize{$\pm$.06}} & {\footnotesize{\bf  0.159}~\scriptsize{$\pm$.01}} & 
    \footnotesize{0.165~\scriptsize{$\pm$.00}} & {\footnotesize{\bf 0.154}~\scriptsize{$\pm$.00}} & 
    \footnotesize{0.172~\scriptsize{$\pm$.01}} & {\footnotesize{\cellcolor{green!10}\bf 0.154}~\scriptsize{$\pm$.00}} \\

    &$2$k   &
    \footnotesize{0.167~\scriptsize{$\pm$.00}} & {\cellcolor{green!10}\footnotesize{\bf 0.156}~\scriptsize{$\pm$.00}} & 
    \footnotesize{0.197~\scriptsize{$\pm$.03}} & {\footnotesize{\bf 0.159}~\scriptsize{$\pm$.00}} & 
    \footnotesize{0.167~\scriptsize{$\pm$.00}} & {\footnotesize{\bf 0.156}~\scriptsize{$\pm$.00}} & 
    \footnotesize{0.222~\scriptsize{$\pm$.05}} & {\footnotesize{\bf 0.157}~\scriptsize{$\pm$.00}} \\
    
    &$2.5$k &
    \footnotesize{0.297~\scriptsize{$\pm$.00}} & {\footnotesize{\bf 0.153}~\scriptsize{$\pm$.00}} & 
    \footnotesize{0.390~\scriptsize{$\pm$.19}} & {\footnotesize{\bf 0.156}~\scriptsize{$\pm$.00}} & 
    \footnotesize{0.297~\scriptsize{$\pm$.00}} & {\footnotesize{\bf 0.153}~\scriptsize{$\pm$.00}} & 
    \footnotesize{0.358~\scriptsize{$\pm$.22}} & {\cellcolor{green!10}\footnotesize{\bf 0.153}~\scriptsize{$\pm$.00}} \\
    
    \bottomrule
    \end{tabularx}
    \end{center}
    \vspace{-5pt}
\end{table*}

\renewcommand{\arraystretch}{1.0}
\setlength{\tabcolsep}{1.5pt}
\begin{table*}[t]
    \caption{{\bf Results using different dimensionality reduction methods.} Using an R-learner, we report the PEHE of our EBM and other benchmark methods over 10 runs (standard deviation in scriptsize): PCA, Feature Agglomeration (FA), Spectral Embedding (SE), Isomap, and KernelPCA (K-PCA) and Autoencoder (AE).}
    \label{tab:res:dim_red_main}
   \vspace{-13pt}
    \begin{center}
    \begin{tabularx}{\textwidth}{l r @{\hskip 2pt} | *{6}{C} >{\columncolor[gray]{.95}}C}
    
    \toprule
   \multicolumn{2}{r|}{Methods}  & {\bf PCA} & {\bf FA} & {\bf SE} & {\bf Isomap} & {\bf K-PCA} & {\bf AE} & {\bf EBM}\\
    
    \midrule
    &$n$& 
    \multicolumn{7}{c}{\it Twins ($d=48$)  with increasing sample size}\\
    \midrule
    
    &$500$ &             
    {\footnotesize{1.092}~\scriptsize{$\pm$.11}} & 
    \footnotesize{1.758~\scriptsize{$\pm$1.1}} & {\footnotesize{1.011}~\scriptsize{$\pm$.00}} & 
    \footnotesize{1.006~\scriptsize{$\pm$.00}} & {\footnotesize{1.015}~\scriptsize{$\pm$.00}} &
    {\footnotesize{0.580}~\scriptsize{$\pm$.03}} &
     \footnotesize{\bf 0.145}~\scriptsize{$\pm$.00} \\
    
    &$1$k &
    {\footnotesize{1.015}~\scriptsize{$\pm$.00}} & 
    \footnotesize{0.963~\scriptsize{$\pm$.00}} & {\footnotesize{1.010}~\scriptsize{$\pm$.00}} & 
    \footnotesize{1.004~\scriptsize{$\pm$.00}} & {\footnotesize{1.010}~\scriptsize{$\pm$.00}} & 
    {\footnotesize{0.549}~\scriptsize{$\pm$.04}} &
    \footnotesize{\bf 0.161}~\scriptsize{$\pm$.00}\\
    
    &$1.5$k&
    \footnotesize{1.014~\scriptsize{$\pm$.00}} & {\footnotesize{0.965}~\scriptsize{$\pm$.00}} & 
    \footnotesize{1.005~\scriptsize{$\pm$.00}} & {\footnotesize{1.006}~\scriptsize{$\pm$.00}} & 
    \footnotesize{1.012~\scriptsize{$\pm$.00}} & 
    {\footnotesize{0.546}~\scriptsize{$\pm$.04}} &
    {\footnotesize{\bf 0.154}~\scriptsize{$\pm$.00}} \\

    &$2$k   &
    \footnotesize{1.013~\scriptsize{$\pm$.00}} & {\footnotesize{0.957}~\scriptsize{$\pm$.00}} & 
    \footnotesize{1.009~\scriptsize{$\pm$.00}} & {\footnotesize{1.007}~\scriptsize{$\pm$.00}} & 
    \footnotesize{1.013~\scriptsize{$\pm$.00}} & 
    {\footnotesize{0.579}~\scriptsize{$\pm$.03}} & 
    {\footnotesize{\bf 0.157}~\scriptsize{$\pm$.00}} \\
    
    &$2.5$k &
    \footnotesize{1.007~\scriptsize{$\pm$.00}} & {\footnotesize{0.951}~\scriptsize{$\pm$.00}} & 
    \footnotesize{1.002~\scriptsize{$\pm$.00}} & {\footnotesize{1.006}~\scriptsize{$\pm$.00}} & 
    \footnotesize{1.006~\scriptsize{$\pm$.00}} & 
    {\footnotesize{0.542}~\scriptsize{$\pm$.04}}& {\footnotesize{\bf 0.153}~\scriptsize{$\pm$.00}} \\
    
    \bottomrule
    \end{tabularx}
    \end{center}
\end{table*}

\subsection{CATE estimation}
Our main contribution is a way to increase performance for {\it any} learner. Specifically, in high dimensions and small sample sizes. 
We evaluate learners' performance using  {\it precision of estimating heterogeneous effects} (PEHE) introduced in \citet{hill2011bayesian} and now standard in CATE estimation. PEHE is essentially the expected risk 
$\mathbb{E} \left\{ [\hat{\tau}(X) - \tau(X)]^2 \right\}$ we define in \Cref{sect:setup}. Because any individual's  treated and control outcomes are never observed jointly, CATEs are unobserved in any real-world data. The literature thus relies on (semi-)synthetic data to evaluate CATE learners.

In our synthetic setup, the generating process of the observed variables $O = (X,A,Y)$ starts by sampling a latent variable $U \sim \mathcal{N}(0, I_{5\times 5})$. Then we  generate a set of covariates $X = \mathcal{N}( g(U),I_{d\times d})$, two potential outcomes, $\mu_0(U)$ and $\mu_1(U)$ and a treatment assignment $A\sim \text{Ber}[\pi(U)]$.  The observed outcome is given by $Y=\mathcal{N}(A\mu_0(U)+(1-A)\mu_1(U),1)$. The CATE  is given by $\tau(U) = \mu_1(U)-\mu_0(U)$. The function $g$ is a deep ReLU network; $\mu_0$ and $\mu_1$ are one-layer neural networks, with an $\exp$-function on their output layers; $\pi$ is a one-layer network with a $\mathop{\mathrm{sigmoid}}$-function on its output layer. By generating i.i.d samples from this process, we create a training set  (with size $n$ specified in \Cref{tab:res:CATE:synth}) and a large testing set with 20k samples.

Given a training set, we first use it to optimize our partially randomized EBM. Then we preprocess it and apply various CATE learners on these lower-dimensional representations. As a comparison, we also apply the same CATE learners on the original covariates. 

\textbf{Lower PEHE across CATE learners.}
Table~\ref{tab:res:CATE:synth} shows that our method greatly benefits a broad spectrum of CATE learners on the synthetic dataset and semi-synthetic dataset Twins \citep{almond2005costs}  with real covariates, especially in small sample sizes. While the gain of using our method diminishes somewhat in larger sample sizes, it is still significant. More importantly, we observe that with our EBM, the performance gaps between different learners shrink significantly. Specifically, R-learner with EBM has the best performance on average over the table while it performs poorly in small samples without EBM. Overall, our experimental results align with our theoretical discussion in \Cref{sect:setup}: by reducing the dimensionality $d$ to a smaller number, the learners will have lower expected errors, i.e., lower PEHEs and smaller performance gaps.

\textbf{Lower PEHE than benchmark dimensionality reduction methods.} Based on our previous experiment, a logical next question to ask is whether other dimensionality reduction methods may also help. We compare our EBM method to various linear and nonlinear dimensionality reduction methods in prepossessing the real covariates of the Twins dataset.  Specifically, we compare against: Principal Components Analysis (PCA), Feature Agglomeration (FA), Spectral Embedding (SE), Isomap, KernelPCA with an RBF kernel, and an Autoencoder (AE).  \Cref{tab:res:dim_red_main} shows that our EBM method outperforms all the benchmarks significantly over different sample sizes.

To further validate our proposed method, we repeat the same experiment using \textbf{additional regression models and data}, and report consistent results to those we present in this section, in \Cref{sect:hyper}.  Overall, we do not found \textbf{sample splitting} increase the variance of our method across all our experiments. As we explained below \cref{equ:rank}, the representation model $f_{\theta}$ is trained with all the samples in our objective function.

\subsection{Partial identifiability of representations}\label{sect:6.2}


In this section we empirically validate that our method produces identifiable representations. Having an identifiable method is important for later inspection of the representations, but also to produce consistent CATE learners. Both of which are important in practice.

\begin{figure}[h]
     {\centering
     \begin{subfigure}[b]{0.43\textwidth}
         \centering
         \includegraphics[width=\textwidth]{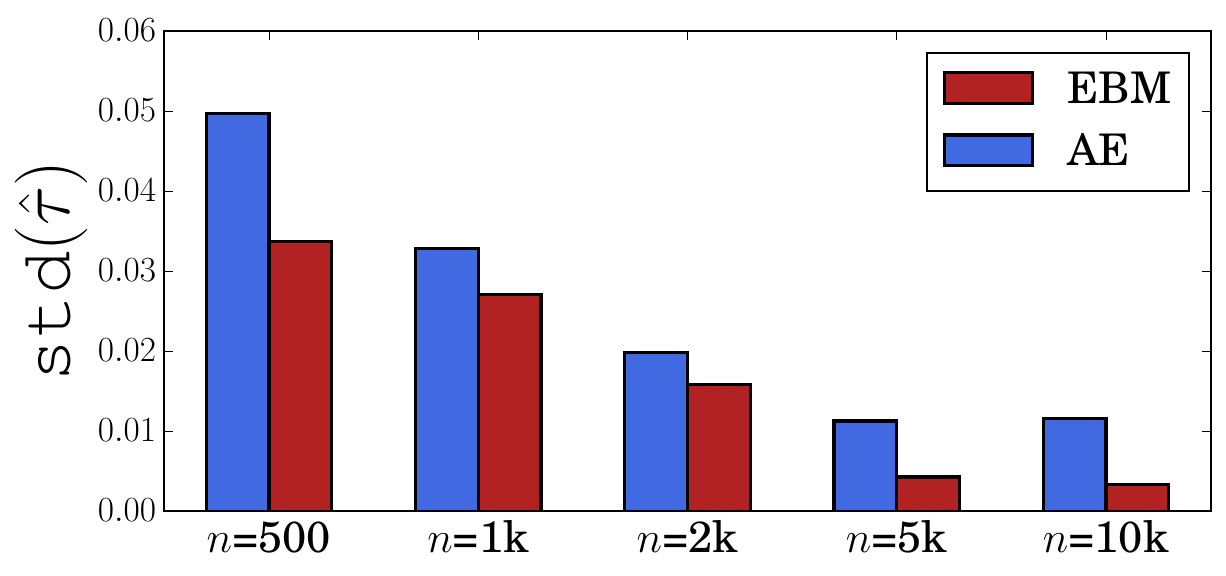}
     \end{subfigure}
     \hfill
     \begin{subfigure}[b]{0.43\textwidth}
         \centering
         \includegraphics[width=\textwidth]{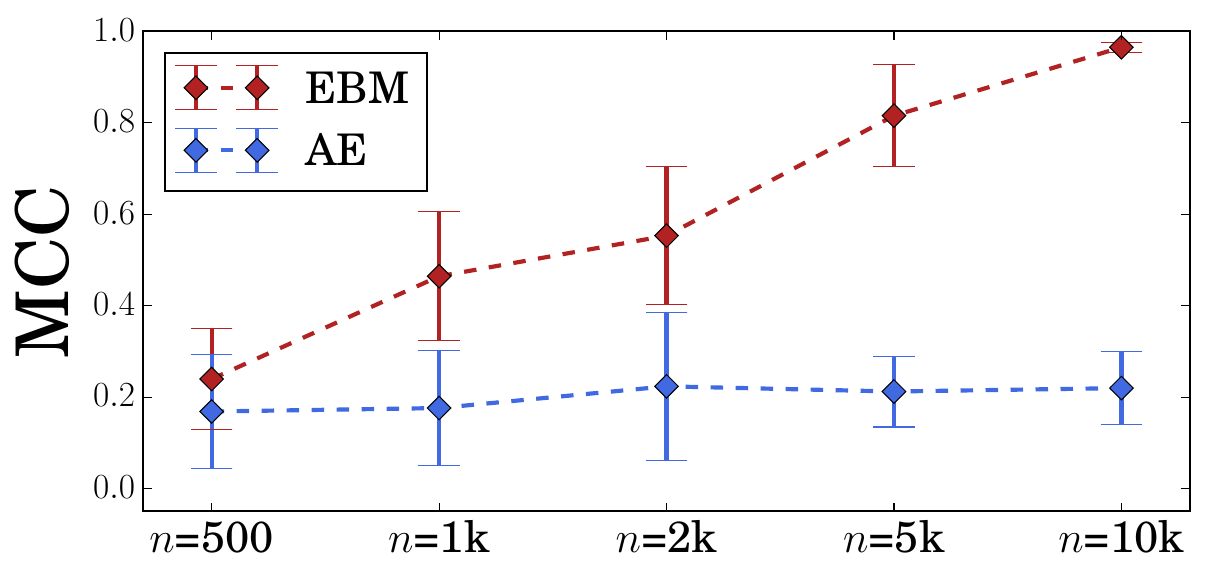}
     \end{subfigure}
    \caption{{\bf Results on identifiability.} {\it Above---} For each model (an autoencoder (AE), and our model (EBM)) we learn ten distinct representations. We then fit an R-Learner on each representation, and calculate the standard deviation of their CATE estimates. Our method has lower standard errors compared to AE. {\it Below---} We report the mean correlation coefficient (MCC) between the representations on the Twins data (higher is better). Our EBM becomes more consistent with larger samples (error bars indicate standard deviation on MCC), and even tends to 1 in large samples.}
    \label{fig:three graphs}}
    \vspace{-5pt}
    \rule{.5\linewidth}{.5pt}
\end{figure}

\textbf{Converging CATE estimates.}
The first panel in \Cref{fig:three graphs} reports the standard deviation of the CATE-estimates, by an R-learner when fitted on the representations of: an autoencoder (AE) and our method (EBM). The representations have the same amount of dimensions ($k=5$). \Cref{fig:three graphs} shows that our model decreases the standard deviation with increasing sample size---this is important, as many applications require estimates to be consistent.

\textbf{Converging representations.}
As discussed at the end of \Cref{sect:method}, the learnt  representations after standardization should correlate as sample size increases. We train our EBM ten times using distinct random initializations, while keeping the random matrix fixed across runs. We subsequently compute the mean correlation coefficient (MCC) between the representations of the test-set from different runs. The MCC is computed by averaging the correlation between {\it each dimension} in the representations of $20$k samples from the test set. Note that the latter is a strict definition as it requires the representation to be consistent for each {\it individual} dimension.\footnote{Previous work \citep{khemakhem2020ice} tests identifiability using the MCC maximized by canonical-correlation analysis (CCA). Here we compute the exact correlation to test our stronger version of identifiability.} Reported in the second panel, we see that our EBM's MCC grows as the sample size increases, leaving the (unidentifiable) AE behind, indicating that our EBM is identifiable, further confirming our theory.

\section{Conclusions}\label{sect:discuss}

We propose a partially randomized EBM with universal approximation capability to learn a partially identifiable low-dimensional representation of moderate or high-dimensional covariates in CATE estimation. We show theoretically and empirically that 
by training our EBM with a noise contrastive loss function and a sample splitting strategy,
our representations converge to a set of limits differing only by some constants. This enables downstream learners to achieve consistent CATE estimates. Experiments on multiple datasets with various dimensions and sample sizes verify our theories and demonstrate a significant performance increase when using our method for CATE estimation.

Our work opens a few new directions for future research. First, our method currently operates within the standard setup of observational data in causal inference, while our partial identifiability theory does not rely on the network architecture of $f_{\theta}$. Extending our work to other high-dimensional settings such as time series or vision could prove useful for many real world applications. Second, as an interpretable approach within CATE estimation, matching is concerned with finding similar individuals across treatment and control groups. While effective, matching becomes harder in high-dimensions. Extending our approach to remain interpretable (e.g. by measuring each covariate's influence to each dimension in the representation) can arm matching approaches against the dimensionality curse. In essence, we believe our method could benefit a wide range of applications requiring causal inference.

\section{Acknowledgements}

This work was supported by GlaxoSmithKline (GSK), the W.D. Armstrong Trust, the National Science Foundation (NSF) under grant number 1722516, the Office of Naval Research (ONR), and The Alan Turning Institute (ATI). We thank all reviewers for their invaluable comments and suggestions.


\balance
\bibliography{ref}
\bibliographystyle{apalike}

\clearpage
\nobalance

\newpage

\appendix
\onecolumn

\section{Additional experiments \& Hyperparameters}\label{sect:hyper}
In \Cref{sect:app:dim_red} we compare a variety of dimensionality reduction methods to our EBM, by using each as a preprocessing step before constructing a CATE estimator. In \Cref{sect:additonal:ML} we repeat our experiments in \Cref{tab:res:CATE:synth} using different regression models to estimate the outcomes and propensity score in the same CATE learners, and using an additional real-world dataset. For hyperparameter settings we refer to \Cref{sect:app:hyper_details}, and for details on used CATE learners we refer to \Cref{sect:app:cate_details}. 

\subsection{Comparision to alternative dimensionality reduction methods} \label{sect:app:dim_red}
In \Cref{tab:res:dim_red} we compare our EBM to a variety of alternative dimensionality reduction methods. Specifically, we compare against: principal components analysis (PCA), Feature Agglomeration (FA), Spectral Embedding (SE), Isomap, KernelPCA with an RBF kernel, and Autoencoder (AE). From \Cref{tab:res:dim_red} we note that none of these methods succeeds in successfully learning informative representations, in such a way to keep downstream CATE learners accurate; our EBM outperforms the benchmarks in most cases over different datasets with various dimensions and sample sizes.

\renewcommand{\arraystretch}{0.5}
\setlength{\tabcolsep}{1.5pt}
\begin{table*}[h]
    \caption{{\bf Results using different dimensionality reduction methods (Copy of \Cref{tab:res:dim_red_main} with additional data).} Using an R-learner, we report the PEHE of our EBM and other benchmark methods: PCA, Feature Agglomeration (FA), Spectral Embedding (SE), Isomap, and KernelPCA (K-PCA) and Autoencoder (AE).}
    \label{tab:res:dim_red}
   \vspace{-13pt}
    \begin{center}
    \begin{tabularx}{\textwidth}{l r @{\hskip 2pt} | *{6}{C} >{\columncolor[gray]{.95}}C}
    
    \toprule
   \multicolumn{2}{r|}{Methods}  & {\bf PCA} & {\bf FA} & {\bf SE} & {\bf Isomap} & {\bf K-PCA} & {\bf AE} & {\bf EBM}\\
  \midrule     
    
      $d$ & $n$& 
    \multicolumn{7}{c}{\it Synth. data with increasing sample size and increasing dimensions}\\
     \midrule   
    
    $50$ &$100$ & 
    \footnotesize{2.139~\scriptsize{$\pm$.00}} & {\footnotesize{2.123}~\scriptsize{$\pm$.00}} & 
    \footnotesize{2.183~\scriptsize{$\pm$.00}} & {\footnotesize{2.135}~\scriptsize{$\pm$.00}} & 
    \footnotesize{2.141~\scriptsize{$\pm$.00}} & 
    {\footnotesize{2.259}~\scriptsize{$\pm$.02}} &{\footnotesize{\bf 1.982}~\scriptsize{$\pm$.01}} \\
    
    $100$ &$250$ & 
    \footnotesize{2.238~\scriptsize{$\pm$.00}} & {\footnotesize{2.239}~\scriptsize{$\pm$.00}} & 
    \footnotesize{2.151~\scriptsize{$\pm$.00}} & {\footnotesize{ 2.231}~\scriptsize{$\pm$.00}} & 
    \footnotesize{2.236~\scriptsize{$\pm$.00}} & 
    {\footnotesize{ 2.055}~\scriptsize{$\pm$.01}} &{\footnotesize{\bf 2.032}~\scriptsize{$\pm$.01}} \\
    
     $150$ &$500$ &
    \footnotesize{2.224~\scriptsize{$\pm$.00}} & {\footnotesize{2.209}~\scriptsize{$\pm$.00}} & 
    \footnotesize{2.963~\scriptsize{$\pm$.00}} & {\footnotesize{2.231}~\scriptsize{$\pm$.00}} & 
    \footnotesize{2.239~\scriptsize{$\pm$.00}} & 
    {\footnotesize{2.092}~\scriptsize{$\pm$.03}} &{\footnotesize{\bf 2.034}~\scriptsize{$\pm$.02}} \\
    
    $200$ &    $1$k & 
    \footnotesize{2.152~\scriptsize{$\pm$.00}} & {\footnotesize{ 2.154}~\scriptsize{$\pm$.00}} & 
    \footnotesize{2.097~\scriptsize{$\pm$.00}} & {\footnotesize{2.168}~\scriptsize{$\pm$.00}} & 
    \footnotesize{2.168~\scriptsize{$\pm$.00}} & 
    {\footnotesize{1.995}~\scriptsize{$\pm$.02}} &{\footnotesize{\bf 1.945}~\scriptsize{$\pm$.01}} \\
    
     $250$ &$1.5$k &
    \footnotesize{2.158~\scriptsize{$\pm$.01}} & {\footnotesize{2.163}~\scriptsize{$\pm$.00}} & 
    \footnotesize{2.401~\scriptsize{$\pm$.00}} & {\footnotesize{2.229}~\scriptsize{$\pm$.00}} & 
    \footnotesize{2.196~\scriptsize{$\pm$.00}} & 
    {\footnotesize{2.071}~\scriptsize{$\pm$.05}} &{\footnotesize{\bf 1.962}~\scriptsize{$\pm$.02}} \\
    
    \midrule
    &$n$& 
    \multicolumn{7}{c}{\it Synth. data with increasing sample size and dimensions fixed at $d=100$}\\
    \midrule
    
    &$100$& 
    \footnotesize{2.194~\scriptsize{$\pm$.03}} & {\footnotesize{2.203}~\scriptsize{$\pm$.04}} & 
    \footnotesize{2.262~\scriptsize{$\pm$.00}} & {\footnotesize{2.156}~\scriptsize{$\pm$.01}} & 
    \footnotesize{2.127~\scriptsize{$\pm$.00}} & 
    {\footnotesize{2.476}~\scriptsize{$\pm$.18}} &{\footnotesize{\bf 1.955}~\scriptsize{$\pm$.03}} \\
    
    &$250$ & 
    \footnotesize{2.238~\scriptsize{$\pm$.00}} & {\footnotesize{2.244}~\scriptsize{$\pm$.01}} & 
    \footnotesize{2.150~\scriptsize{$\pm$.00}} & {\footnotesize{2.233}~\scriptsize{$\pm$.01}} & 
    \footnotesize{2.232~\scriptsize{$\pm$.00}} &
    {\footnotesize{2.109}~\scriptsize{$\pm$.02}} &{\footnotesize{\bf 2.032}~\scriptsize{$\pm$.01}} \\
     
    &$500$ &
    \footnotesize{2.119~\scriptsize{$\pm$.01}} & {\footnotesize{2.029}~\scriptsize{$\pm$.01}} & 
    \footnotesize{2.395~\scriptsize{$\pm$.00}} & {\footnotesize{2.230}~\scriptsize{$\pm$.00}} & 
    \footnotesize{2.118~\scriptsize{$\pm$.00}} & 
    {\footnotesize{2.092}~\scriptsize{$\pm$.04}} &{\footnotesize{\bf 2.008}~\scriptsize{$\pm$.00}} \\

    &$1$k  &
    \footnotesize{2.183~\scriptsize{$\pm$.00}} & {\footnotesize{2.231}~\scriptsize{$\pm$.00}} & 
    \footnotesize{2.145~\scriptsize{$\pm$.00}} & {\footnotesize{2.219}~\scriptsize{$\pm$.00}} & 
    \footnotesize{2.199~\scriptsize{$\pm$.00}} & 
    {\footnotesize{2.059}~\scriptsize{$\pm$.01}} &{\footnotesize{\bf 1.987}~\scriptsize{$\pm$.02}} \\

    &$1.5$k &
    \footnotesize{2.174~\scriptsize{$\pm$.00}} & {\footnotesize{2.213}~\scriptsize{$\pm$.00}} & 
    \footnotesize{2.083~\scriptsize{$\pm$.00}} & {\footnotesize{2.229}~\scriptsize{$\pm$.00}} & 
    \footnotesize{2.199~\scriptsize{$\pm$.00}} & 
    {\footnotesize{2.048}~\scriptsize{$\pm$.01}} &{\footnotesize{\bf 2.007}~\scriptsize{$\pm$.01}} \\

    \midrule
    &$n$& 
    \multicolumn{7}{c}{\it Twins ($d=48$)  with increasing sample size}\\
    \midrule
    
    &$500$ &             
    {\footnotesize{1.092}~\scriptsize{$\pm$.11}} & 
    \footnotesize{1.758~\scriptsize{$\pm$1.1}} & {\footnotesize{1.011}~\scriptsize{$\pm$.00}} & 
    \footnotesize{1.006~\scriptsize{$\pm$.00}} & {\footnotesize{1.015}~\scriptsize{$\pm$.00}} &
    {\footnotesize{0.580}~\scriptsize{$\pm$.03}} &
     \footnotesize{\bf 0.145}~\scriptsize{$\pm$.00} \\
    
    &$1$k &
    {\footnotesize{1.015}~\scriptsize{$\pm$.00}} & 
    \footnotesize{0.963~\scriptsize{$\pm$.00}} & {\footnotesize{1.010}~\scriptsize{$\pm$.00}} & 
    \footnotesize{1.004~\scriptsize{$\pm$.00}} & {\footnotesize{1.010}~\scriptsize{$\pm$.00}} & 
    {\footnotesize{0.549}~\scriptsize{$\pm$.04}} &
    \footnotesize{\bf 0.161}~\scriptsize{$\pm$.00}\\
    
    &$1.5$k&
    \footnotesize{1.014~\scriptsize{$\pm$.00}} & {\footnotesize{0.965}~\scriptsize{$\pm$.00}} & 
    \footnotesize{1.005~\scriptsize{$\pm$.00}} & {\footnotesize{1.006}~\scriptsize{$\pm$.00}} & 
    \footnotesize{1.012~\scriptsize{$\pm$.00}} & 
    {\footnotesize{0.546}~\scriptsize{$\pm$.04}} &
    {\footnotesize{\bf 0.154}~\scriptsize{$\pm$.00}} \\

    &$2$k   &
    \footnotesize{1.013~\scriptsize{$\pm$.00}} & {\footnotesize{0.957}~\scriptsize{$\pm$.00}} & 
    \footnotesize{1.009~\scriptsize{$\pm$.00}} & {\footnotesize{1.007}~\scriptsize{$\pm$.00}} & 
    \footnotesize{1.013~\scriptsize{$\pm$.00}} & 
    {\footnotesize{0.579}~\scriptsize{$\pm$.03}} & 
    {\footnotesize{\bf 0.157}~\scriptsize{$\pm$.00}} \\
    
    &$2.5$k &
    \footnotesize{1.007~\scriptsize{$\pm$.00}} & {\footnotesize{0.951}~\scriptsize{$\pm$.00}} & 
    \footnotesize{1.002~\scriptsize{$\pm$.00}} & {\footnotesize{1.006}~\scriptsize{$\pm$.00}} & 
    \footnotesize{1.006~\scriptsize{$\pm$.00}} & 
    {\footnotesize{0.542}~\scriptsize{$\pm$.04}} &{\footnotesize{\bf 0.153}~\scriptsize{$\pm$.00}} \\
    
    \midrule
    &$n$& 
    \multicolumn{7}{c}{\it IHDP ($d=25$)  with increasing sample size}\\
    \midrule
    
    &$100$ &             
    \footnotesize{6.266~\scriptsize{$\pm$.34}} & {\footnotesize{\bf 1.833}~\scriptsize{$\pm$.02}} & 
    \footnotesize{5.295~\scriptsize{$\pm$.00}} & {\footnotesize{5.276}~\scriptsize{$\pm$.02}} & 
    \footnotesize{5.448~\scriptsize{$\pm$.04}} & 
    {\footnotesize{3.580}~\scriptsize{$\pm$1.2}} &{\footnotesize{2.444}~\scriptsize{$\pm$.73}} \\
    
    &$250$ &
    \footnotesize{6.243~\scriptsize{$\pm$.06}} & {\footnotesize{\bf 1.624}~\scriptsize{$\pm$.04}} & 
    \footnotesize{5.180~\scriptsize{$\pm$.00}} & {\footnotesize{6.145}~\scriptsize{$\pm$.07}} & 
    \footnotesize{5.564~\scriptsize{$\pm$.02}} & 
    {\footnotesize{2.811}~\scriptsize{$\pm$.73}} &{\footnotesize{1.729}~\scriptsize{$\pm$.19}} \\
    
    &$500$&
    \footnotesize{6.995~\scriptsize{$\pm$.05}} & {\footnotesize{1.951}~\scriptsize{$\pm$.01}} & 
    \footnotesize{5.293~\scriptsize{$\pm$.00}} & {\footnotesize{5.988}~\scriptsize{$\pm$.06}} & 
    \footnotesize{6.469~\scriptsize{$\pm$.15}} & 
    {\footnotesize{2.671}~\scriptsize{$\pm$.14}} &{\footnotesize{\bf 1.635}~\scriptsize{$\pm$.09}} \\

    \bottomrule
    \end{tabularx}
    \end{center}
\end{table*}

\subsection{CATE learners with different regression models, and different data} \label{sect:additonal:ML}
Consider Tables \ref{tab:res:CATE:extra:pow}-\ref{tab:res:CATE:extra:poly}-\ref{tab:res:CATE:extra:lin}, where we report the PEHE given the same experimental setup as we have in \Cref{tab:res:CATE:synth}; for additional data (Infant Health Development Program (IHDP) \citep{macdorman1999infant}), and three additional regression models (\texttt{PowerTransform Regression} \citep{yeo2000new}, \texttt{Polynomial Regression}, and \texttt{Ridge Regression}, respectively). From our results we learn that our EBM is agnostic to the choice of regression model, and is versatile enough to also perform well given other data. These results are promising and should give some assurance regarding our method before application in practice. As we have in \Cref{tab:res:CATE:synth}, we ran each CATE learner on ten distinct representations, given different folds of the data, and averaged the results. Note that we have not specifically optimised the EBM's hyperparameters for these different regression models, but rather kept them as they were in \Cref{tab:res:CATE:synth} (actual hyperparameter values are reported in \Cref{tab:hyperparams}).
We also include a \quotes{complete} version of \Cref{tab:res:CATE:synth} in \Cref{tab:res:CATE:ihdp_main}, where we include results on IHDP as an additional dataset. Note that these results are in line with those reported earlier using different regression models.

\renewcommand{\arraystretch}{0.5}
\setlength{\tabcolsep}{1.5pt}
\begin{table*}[t]
    \caption{{\bf Results on (semi-)synthetic data (Twins \& IHDP) with \texttt{PowerTransform Regression}.} We report for the same configuration as in \Cref{tab:res:CATE:synth}. Results are averaged over ten runs with (\quotes{\cmark}), and without (\quotes{\xmark}) the same representations used in \Cref{tab:res:CATE:synth}.}
       \vspace{-13pt}
    \label{tab:res:CATE:extra:pow}
\begin{center}
    \begin{tabularx}{\textwidth}{l r @{\hskip 2pt}  *{4}{|CC}}
    
    \toprule
    \multicolumn{2}{r|}{Methods}   
        & \multicolumn{2}{c}{\bf X-Learner}  
        & \multicolumn{2}{c}{\bf DR-Learner} 
        & \multicolumn{2}{c}{\bf T-Learner} 
        & \multicolumn{2}{c}{\bf R-Learner}\\
    \midrule     
    \multicolumn{2}{r|}{EBM} & \xmark & \cmark & \xmark & \cmark & \xmark & \cmark & \xmark & \cmark \\
    
    \midrule  
    
      $d$ & $n$& 
    \multicolumn{8}{c}{\it Synth. data with increasing sample size and increasing dimensions}\\
     \midrule   
    
    $50$ &$100$ & 
    \footnotesize{2.267~\scriptsize{$\pm$.00}} & {\cellcolor{green!10}\footnotesize{\bf 2.010}~\scriptsize{$\pm$.03}} & 
    \footnotesize{5.593~\scriptsize{$\pm$2.2}} & {\footnotesize{\bf 2.015}~\scriptsize{$\pm$.05}} & 
    \footnotesize{2.455~\scriptsize{$\pm$.00}} & {\footnotesize{\bf 2.011}~\scriptsize{$\pm$.03}} &  
    \footnotesize{53.17~\scriptsize{$\pm$5.2}} & {\footnotesize{\bf 11.16}~\scriptsize{$\pm$3.9}}  \\
    
    $100$ &$250$ & 
    \footnotesize{2.754~\scriptsize{$\pm$.00}} & {\cellcolor{green!10}\footnotesize{\bf 2.019}~\scriptsize{$\pm$.01}} & 
    \footnotesize{3.963~\scriptsize{$\pm$.24}} & {\footnotesize{\bf 2.027}~\scriptsize{$\pm$1.2}} & 
    \footnotesize{2.798~\scriptsize{$\pm$.00}} & {\footnotesize{\bf 2.020}~\scriptsize{$\pm$.01}} &  
    \footnotesize{60.10~\scriptsize{$\pm$4.2}} & {\footnotesize{\bf 10.70}~\scriptsize{$\pm$3.9}}  \\
    
     $150$ &$500$ &
    \footnotesize{2.575~\scriptsize{$\pm$.00}} & {\footnotesize{\bf 2.002}~\scriptsize{$\pm$.01}} & 
    \footnotesize{2.986~\scriptsize{$\pm$.08}} & {\cellcolor{green!10}\footnotesize{\bf 2.001}~\scriptsize{$\pm$.01}} & 
    \footnotesize{2.595~\scriptsize{$\pm$.00}} & {\footnotesize{\bf 2.001}~\scriptsize{$\pm$.01}} &  
    \footnotesize{49.78~\scriptsize{$\pm$4.2}} & {\footnotesize{\bf 12.31}~\scriptsize{$\pm$1.8}}  \\
    
    $200$ &    $1$k & 
    \footnotesize{2.197~\scriptsize{$\pm$.00}} & {\footnotesize{\bf 1.952}~\scriptsize{$\pm$.00}} & 
    \footnotesize{2.293~\scriptsize{$\pm$.03}} & {\cellcolor{green!10}\footnotesize{\bf 1.941}~\scriptsize{$\pm$.01}} & 
    \footnotesize{2.202~\scriptsize{$\pm$.00}} & {\footnotesize{\bf 1.951}~\scriptsize{$\pm$.01}} &  
    \footnotesize{42.76~\scriptsize{$\pm$3.9}} & {\footnotesize{\bf 2.838}~\scriptsize{$\pm$.64}}  \\
    
     $250$ &$1.5$k &
    \footnotesize{2.288~\scriptsize{$\pm$.00}} & {\footnotesize{\bf 1.966}~\scriptsize{$\pm$.04}} & 
    \footnotesize{2.410~\scriptsize{$\pm$.04}} & {\footnotesize{\bf 1.979}~\scriptsize{$\pm$.03}} & 
    \footnotesize{2.295~\scriptsize{$\pm$.00}} & {\cellcolor{green!10}\footnotesize{\bf 1.968}~\scriptsize{$\pm$.04}} &  
    \footnotesize{42.03~\scriptsize{$\pm$3.6}} & {\footnotesize{\bf 2.535}~\scriptsize{$\pm$.15}}  \\
    
    \midrule
    &$n$& 
    \multicolumn{8}{c}{\it Synth. data with increasing sample size and dimensions fixed at $d=100$}\\
    \midrule
    
     &$100$& 
    \footnotesize{2.150~\scriptsize{$\pm$.00}} & {\cellcolor{green!10}\footnotesize{\bf 1.964}~\scriptsize{$\pm$.02}} & 
    \footnotesize{32.63~\scriptsize{$\pm$13}} & {\footnotesize{\bf 2.147}~\scriptsize{$\pm$.15}} & 
    \footnotesize{2.289~\scriptsize{$\pm$.00}} & {\footnotesize{\bf 1.973}~\scriptsize{$\pm$.03}} &  
    \footnotesize{58.80~\scriptsize{$\pm$5.1}} & {\footnotesize{\bf 5.713}~\scriptsize{$\pm$2.2}}  \\
    
    &$250$ & 
    \footnotesize{2.754~\scriptsize{$\pm$.00}} & {\cellcolor{green!10}\footnotesize{\bf 2.019}~\scriptsize{$\pm$.01}} & 
    \footnotesize{3.963~\scriptsize{$\pm$.24}} & {\footnotesize{\bf 2.027}~\scriptsize{$\pm$1.2}} & 
    \footnotesize{2.798~\scriptsize{$\pm$.00}} & {\footnotesize{\bf 2.020}~\scriptsize{$\pm$.01}} &  
    \footnotesize{60.10~\scriptsize{$\pm$4.2}} & {\footnotesize{\bf 10.70}~\scriptsize{$\pm$3.9}}  \\
     
    &$500$ &
    \footnotesize{2.150~\scriptsize{$\pm$.00}} & {\footnotesize{\bf 2.029}~\scriptsize{$\pm$.02}} & 
    \footnotesize{2.319~\scriptsize{$\pm$.06}} & {\cellcolor{green!10}\footnotesize{\bf 2.006}~\scriptsize{$\pm$.05}} & 
    \footnotesize{2.160~\scriptsize{$\pm$.00}} & {\footnotesize{\bf 2.028}~\scriptsize{$\pm$.03}} &  
    \footnotesize{41.85~\scriptsize{$\pm$3.4}} & {\footnotesize{\bf 3.884}~\scriptsize{$\pm$.79}}  \\

    &$1$k  &
    \footnotesize{2.053~\scriptsize{$\pm$.00}} & {\cellcolor{green!10}\footnotesize{\bf 1.986}~\scriptsize{$\pm$.02}} & 
    \footnotesize{2.102~\scriptsize{$\pm$.01}} & {\footnotesize{\bf 1.989}~\scriptsize{$\pm$.01}} & 
    \footnotesize{2.057~\scriptsize{$\pm$.00}} & {\footnotesize{\bf 1.987}~\scriptsize{$\pm$.02}} &  
    \footnotesize{39.49~\scriptsize{$\pm$3.2}} & {\footnotesize{\bf 2.637}~\scriptsize{$\pm$.20}}  \\

    &$1.5$k &
    \footnotesize{2.008~\scriptsize{$\pm$.00}} & {\cellcolor{green!10}\footnotesize{\bf 1.999}~\scriptsize{$\pm$.02}} & 
    \footnotesize{2.354}~\scriptsize{$\pm$.01} & {\footnotesize{\bf 1.999}~\scriptsize{$\pm$.52}} & 
    \footnotesize{2.008~\scriptsize{$\pm$.00}} & {\footnotesize{\bf 2.000}~\scriptsize{$\pm$.02}} &  
    \footnotesize{37.17~\scriptsize{$\pm$2.9}} & {\footnotesize{\bf 3.949}~\scriptsize{$\pm$.77}}  \\

    \midrule
    &$n$& 
    \multicolumn{8}{c}{\it Twins ($d=48$)  with increasing sample size}\\
    \midrule
    
    &$500$ &             
    \footnotesize{0.203~\scriptsize{$\pm$.00}} & {\footnotesize{\bf 0.187}~\scriptsize{$\pm$.06}} & 
    \footnotesize{4.383~\scriptsize{$\pm$.22}} & {\cellcolor{green!10}\footnotesize{\bf 0.185}~\scriptsize{$\pm$.02}} & 
    \footnotesize{\bf 0.204}~\scriptsize{$\pm$.00} & {\footnotesize{0.248}~\scriptsize{$\pm$.24}} &  
    \footnotesize{300.0~\scriptsize{$\pm$16.}} & {\footnotesize{\bf 2.176}~\scriptsize{$\pm$.89}}  \\
    
    &$1$k &
    \footnotesize{0.177~\scriptsize{$\pm$.00}} & {\footnotesize{\bf 0.169}~\scriptsize{$\pm$.03}} & 
    \footnotesize{0.194~\scriptsize{$\pm$.01}} & {\footnotesize{\bf 0.163}~\scriptsize{$\pm$.01}} & 
    \footnotesize{0.177~\scriptsize{$\pm$.00}} & {\cellcolor{green!10}\footnotesize{\bf 0.159}~\scriptsize{$\pm$.02}} &  
    \footnotesize{31.39~\scriptsize{$\pm$2.5}} & {\footnotesize{\bf 1.029}~\scriptsize{$\pm$1.1}}  \\
    
    &$1.5$k&
    \footnotesize{0.169~\scriptsize{$\pm$.00}} & {\cellcolor{green!10}\footnotesize{\bf 0.154}~\scriptsize{$\pm$.00}} & 
    \footnotesize{\bf 0.172}~\scriptsize{$\pm$.01} & {\footnotesize{ 0.183}~\scriptsize{$\pm$.08}} & 
    \footnotesize{0.169~\scriptsize{$\pm$.00}} & {\footnotesize{\bf 0.155}~\scriptsize{$\pm$.00}} &  
    \footnotesize{31.23~\scriptsize{$\pm$2.3}} & {\footnotesize{\bf 0.459}~\scriptsize{$\pm$.15}}  \\

    &$2$k   &
    \footnotesize{0.167~\scriptsize{$\pm$.00}} & {\footnotesize{\bf 0.161}~\scriptsize{$\pm$.00}} & 
    \footnotesize{0.168~\scriptsize{$\pm$.00}} & {\footnotesize{\bf 0.163}~\scriptsize{$\pm$.00}} & 
    \footnotesize{0.168~\scriptsize{$\pm$.00}} & {\cellcolor{green!10}\footnotesize{\bf 0.161}~\scriptsize{$\pm$.00}} &  
    \footnotesize{29.95~\scriptsize{$\pm$2.4}} & {\footnotesize{\bf 0.629}~\scriptsize{$\pm$.30}}  \\
    
    &$2.5$k &
    \footnotesize{0.169~\scriptsize{$\pm$.00}} & {\cellcolor{green!10}\footnotesize{\bf 0.162}~\scriptsize{$\pm$.00}} & 
    \footnotesize{0.170~\scriptsize{$\pm$.00}} & {\footnotesize{\bf 0.163}~\scriptsize{$\pm$.00}} & 
    \footnotesize{0.169~\scriptsize{$\pm$.00}} & {\footnotesize{\bf 0.162}~\scriptsize{$\pm$.00}} &  
    \footnotesize{29.79~\scriptsize{$\pm$2.4}} & {\footnotesize{\bf 0.439}~\scriptsize{$\pm$.19}}  \\
    
    \midrule
    &$n$& 
    \multicolumn{8}{c}{\it IHDP ($d=25$)  with increasing sample size}\\
    \midrule
    
    &$100$ &             
    \footnotesize{1.814~\scriptsize{$\pm$.01}} & {\cellcolor{green!10}\footnotesize{\bf 1.502}~\scriptsize{$\pm$.03}} & 
    \footnotesize{3.755~\scriptsize{$\pm$.72}} & {\footnotesize{\bf 1.637}~\scriptsize{$\pm$.12}} & 
    \footnotesize{1.845~\scriptsize{$\pm$.00}} & {\footnotesize{\bf 1.507}~\scriptsize{$\pm$.05}} &  
    \footnotesize{35.77~\scriptsize{$\pm$6.3}} & {\footnotesize{\bf 21.36}~\scriptsize{$\pm$17.}}  \\
    
    &$250$ &
    \footnotesize{1.713~\scriptsize{$\pm$.00}} & {\footnotesize{\bf 1.598}~\scriptsize{$\pm$.04}} & 
    \footnotesize{1.837~\scriptsize{$\pm$.09}} & {\footnotesize{\bf 1.653}~\scriptsize{$\pm$.13}} & 
    \footnotesize{1.727~\scriptsize{$\pm$.00}} & {\cellcolor{green!10}\footnotesize{\bf 1.593}~\scriptsize{$\pm$.03}} &  
    \footnotesize{11.71~\scriptsize{$\pm$1.5}} & {\footnotesize{\bf 9.552}~\scriptsize{$\pm$4.0}}  \\
    
    &$500$&
    \footnotesize{1.603~\scriptsize{$\pm$.00}} & {\cellcolor{green!10}\footnotesize{\bf 1.554}~\scriptsize{$\pm$.02}} & 
    \footnotesize{1.672~\scriptsize{$\pm$.05}} & {\footnotesize{\bf 1.571}~\scriptsize{$\pm$.03}} & 
    \footnotesize{1.627~\scriptsize{$\pm$.00}} & {\footnotesize{\bf 1.556}~\scriptsize{$\pm$.02}} &  
    \footnotesize{23.45~\scriptsize{$\pm$2.5}} & {\footnotesize{\bf 15.19}~\scriptsize{$\pm$3.9}}  \\
    
    \bottomrule
    \end{tabularx}
    \end{center}
\end{table*}

\renewcommand{\arraystretch}{0.5}
\setlength{\tabcolsep}{1.5pt}
\begin{table*}[b]
    \caption{{\bf Results on (semi-)synthetic data (Twins \& IHDP) using \texttt{Polynomial Regression}.} We report for the same configuration as in \Cref{tab:res:CATE:synth}. Results are averaged over ten runs with (\quotes{\cmark}), and without (\quotes{\xmark}) the same representations used in \Cref{tab:res:CATE:synth}.}
       \vspace{-13pt}
    \label{tab:res:CATE:extra:poly}
    \begin{center}
    \begin{tabularx}{\textwidth}{l r @{\hskip 2pt}  *{4}{|CC}}
    
    \toprule
       \multicolumn{2}{r|}{Methods}   
            & \multicolumn{2}{c}{\bf X-Learner}  
            & \multicolumn{2}{c}{\bf DR-Learner} 
            & \multicolumn{2}{c}{\bf T-Learner} 
            & \multicolumn{2}{c}{\bf R-Learner}\\
      \midrule     
     \multicolumn{2}{r|}{EBM} & \xmark & \cmark & \xmark & \cmark & \xmark & \cmark & \xmark & \cmark \\
    
    \midrule
    
      $d$ & $n$& 
    \multicolumn{8}{c}{\it Synth. data with increasing sample size and increasing dimensions}\\
    \midrule   
    
    $50$ &$100$ & 
    \footnotesize{\bf 2.095}~\scriptsize{$\pm$.00} & {\footnotesize{2.089}~\scriptsize{$\pm$.09}} & 
    \footnotesize{75.12~\scriptsize{$\pm$26.}} & {\footnotesize{\bf 2.500}~\scriptsize{$\pm$.44}} & 
    \footnotesize{2.124~\scriptsize{$\pm$.00}} & {\cellcolor{green!10}\footnotesize{\bf 2.095}~\scriptsize{$\pm$.09}} &  
    \footnotesize{110.8~\scriptsize{$\pm$7.1}} & {\footnotesize{\bf 26.61}~\scriptsize{$\pm$7.6}}  \\
    
    $100$ &$250$ & 
    \footnotesize{2.109~\scriptsize{$\pm$.00}} & {\cellcolor{green!10}\footnotesize{\bf 2.026}~\scriptsize{$\pm$.03}} & 
    \footnotesize{11.67~\scriptsize{$\pm$1.9}} & {\footnotesize{\bf 2.138}~\scriptsize{$\pm$.30}} & 
    \footnotesize{2.168~\scriptsize{$\pm$.00}} & {\footnotesize{\bf 2.028}~\scriptsize{$\pm$.03}} &  
    \footnotesize{50.45~\scriptsize{$\pm$4.4}} & {\footnotesize{\bf 10.72}~\scriptsize{$\pm$4.0}}  \\
    
     $150$ &$500$ &
    \footnotesize{2.048~\scriptsize{$\pm$.00}} & {\footnotesize{\bf 2.008}~\scriptsize{$\pm$.02}} & 
    \footnotesize{8.668~\scriptsize{$\pm$1.5}} & {\cellcolor{green!10}\footnotesize{\bf 2.006}~\scriptsize{$\pm$.01}} & 
    \footnotesize{2.142~\scriptsize{$\pm$.00}} & {\footnotesize{\bf 2.008}~\scriptsize{$\pm$.02}} &  
    \footnotesize{44.29~\scriptsize{$\pm$3.4}} & {\footnotesize{\bf 12.35}~\scriptsize{$\pm$1.8}}  \\
    
    $200$ &    $1$k & 
    \footnotesize{1.964~\scriptsize{$\pm$.00}} & {\footnotesize{\bf 1.949}~\scriptsize{$\pm$.02}} & 
    \footnotesize{7.278~\scriptsize{$\pm$1.3}} & {\cellcolor{green!10}\footnotesize{\bf 1.949}~\scriptsize{$\pm$.02}} & 
    \footnotesize{2.088~\scriptsize{$\pm$.00}} & {\footnotesize{\bf 1.949}~\scriptsize{$\pm$.02}} &  
    \footnotesize{42.04~\scriptsize{$\pm$3.4}} & {\footnotesize{\bf 2.973}~\scriptsize{$\pm$.57}}  \\
    
     $250$ &$1.5$k &
    \footnotesize{1.945~\scriptsize{$\pm$.00}} & {\footnotesize{\bf 1.945}~\scriptsize{$\pm$.03}} & 
    \footnotesize{6.764~\scriptsize{$\pm$1.2}} & {\footnotesize{\bf 1.979}~\scriptsize{$\pm$.03}} & 
    \footnotesize{2.109~\scriptsize{$\pm$.00}} & {\cellcolor{green!10}\footnotesize{\bf 1.945}~\scriptsize{$\pm$.04}} &  
    \footnotesize{41.71~\scriptsize{$\pm$3.5}} & {\footnotesize{\bf 2.555}~\scriptsize{$\pm$.18}}  \\
    
    \midrule
    &$n$& 
    \multicolumn{8}{c}{\it Synth. data with increasing sample size and dimensions fixed at $d=100$}\\
    \midrule
    
     &$100$& 
    \footnotesize{2.009~\scriptsize{$\pm$.00}} & {\cellcolor{green!10}\footnotesize{\bf 1.987}~\scriptsize{$\pm$.09}} & 
    \footnotesize{157.1~\scriptsize{$\pm$50.}} & {\footnotesize{\bf 2.176}~\scriptsize{$\pm$.13}} & 
    \footnotesize{2.038~\scriptsize{$\pm$.00}} & {\footnotesize{\bf 1.992}~\scriptsize{$\pm$.09}} &  
    \footnotesize{54.81~\scriptsize{$\pm$4.3}} & {\footnotesize{\bf 6.612}~\scriptsize{$\pm$3.5}}  \\
    
    &$250$ & 
    \footnotesize{2.109~\scriptsize{$\pm$.00}} & {\cellcolor{green!10}\footnotesize{\bf 2.019}~\scriptsize{$\pm$.01}} & 
    \footnotesize{11.67~\scriptsize{$\pm$1.9}} & {\footnotesize{\bf 2.028}~\scriptsize{$\pm$.02}} & 
    \footnotesize{2.168~\scriptsize{$\pm$.00}} & {\footnotesize{\bf 2.019}~\scriptsize{$\pm$.01}} &  
    \footnotesize{50.46~\scriptsize{$\pm$4.4}} & {\footnotesize{\bf 9.623}~\scriptsize{$\pm$3.0}}  \\
     
    &$500$ & 
    \footnotesize{\cellcolor{green!10}\bf 1.897}~\scriptsize{$\pm$.00} & {\footnotesize{ 2.055}~\scriptsize{$\pm$.05}} & 
    \footnotesize{8.020~\scriptsize{$\pm$1.7}} & {\footnotesize{\bf 2.003}~\scriptsize{$\pm$.00}} & 
    \footnotesize{\bf 2.007}~\scriptsize{$\pm$.00} & {\footnotesize{ 2.051}~\scriptsize{$\pm$.05}} &  
    \footnotesize{43.56~\scriptsize{$\pm$3.9}} & {\footnotesize{\bf 3.758}~\scriptsize{$\pm$.82}}  \\

    &$1$k  &
    \footnotesize{2.210~\scriptsize{$\pm$.00}} & {\cellcolor{green!10}\footnotesize{\bf 1.995}~\scriptsize{$\pm$.02}} & 
    \footnotesize{5.871~\scriptsize{$\pm$.90}} & {\footnotesize{\bf 2.289}~\scriptsize{$\pm$.88}} & 
    \footnotesize{2.338~\scriptsize{$\pm$1.4}} & {\footnotesize{\bf 1.996}~\scriptsize{$\pm$.02}} &  
    \footnotesize{40.38~\scriptsize{$\pm$3.1}} & {\footnotesize{\bf 2.965}~\scriptsize{$\pm$.69}}  \\

    &$1.5$k &
    \footnotesize{2.341~\scriptsize{$\pm$.00}} & {\footnotesize{\bf 2.002}~\scriptsize{$\pm$.02}} & 
    \footnotesize{4.637~\scriptsize{$\pm$.75}} & {\footnotesize{\bf 2.156}~\scriptsize{$\pm$.47}} & 
    \footnotesize{2.474~\scriptsize{$\pm$.00}} & {\cellcolor{green!10}\footnotesize{\bf 2.000}~\scriptsize{$\pm$.02}} &  
    \footnotesize{38.31~\scriptsize{$\pm$2.9}} & {\footnotesize{\bf 3.945}~\scriptsize{$\pm$.76}}  \\

    \midrule
    &$n$& 
    \multicolumn{8}{c}{\it Twins ($d=48$)  with increasing sample size}\\
    \midrule
    
    &$500$ &             
    \footnotesize{0.345~\scriptsize{$\pm$.00}} & {\footnotesize{\bf 0.155}~\scriptsize{$\pm$.00}} & 
    \footnotesize{4.538~\scriptsize{$\pm$1.4}} & {\footnotesize{\bf 0.158}~\scriptsize{$\pm$.00}} & 
    \footnotesize{0.377~\scriptsize{$\pm$.00}} & {\cellcolor{green!10}\footnotesize{\bf 0.155}~\scriptsize{$\pm$.00}} &  
    \footnotesize{116.0~\scriptsize{$\pm$14.}} & {\footnotesize{\bf 0.847}~\scriptsize{$\pm$.19}}  \\
    
    &$1$k &
    \footnotesize{0.486~\scriptsize{$\pm$.00}} & {\cellcolor{green!10}\footnotesize{\bf 0.149}~\scriptsize{$\pm$.00}} & 
    \footnotesize{1.747~\scriptsize{$\pm$.33}} & {\footnotesize{\bf 0.157}~\scriptsize{$\pm$.01}} & 
    \footnotesize{0.529~\scriptsize{$\pm$.00}} & {\footnotesize{\bf 0.149}~\scriptsize{$\pm$.00}} &  
    \footnotesize{78.22~\scriptsize{$\pm$7.3}} & {\footnotesize{\bf 0.482}~\scriptsize{$\pm$.11}}  \\
    
    &$1.5$k&
    \footnotesize{0.455~\scriptsize{$\pm$.00}} & {\cellcolor{green!10}\footnotesize{\bf 0.153}~\scriptsize{$\pm$.00}} & 
    \footnotesize{1.453~\scriptsize{$\pm$.40}} & {\footnotesize{\bf 0.186}~\scriptsize{$\pm$.06}} & 
    \footnotesize{0.481~\scriptsize{$\pm$.00}} & {\footnotesize{\bf 0.153}~\scriptsize{$\pm$.00}} &  
    \footnotesize{142.3~\scriptsize{$\pm$21.}} & {\footnotesize{\bf 0.395}~\scriptsize{$\pm$.13}}  \\

    &$2$k   &
    \footnotesize{0.426~\scriptsize{$\pm$.00}} & {\cellcolor{green!10}\footnotesize{\bf 0.159}~\scriptsize{$\pm$.00}} & 
    \footnotesize{1.109~\scriptsize{$\pm$.32}} & {\footnotesize{\bf 0.162}~\scriptsize{$\pm$.00}} & 
    \footnotesize{0.451~\scriptsize{$\pm$.00}} & {\footnotesize{\bf 0.159}~\scriptsize{$\pm$.00}} &  
    \footnotesize{38.13~\scriptsize{$\pm$4.2}} & {\footnotesize{\bf 0.655}~\scriptsize{$\pm$.33}}  \\
    
    &$2.5$k &
    \footnotesize{0.403~\scriptsize{$\pm$.00}} & {\cellcolor{green!10}\footnotesize{\bf 0.159}~\scriptsize{$\pm$.00}} & 
    \footnotesize{0.921~\scriptsize{$\pm$.14}} & {\footnotesize{\bf 0.163}~\scriptsize{$\pm$.01}} & 
    \footnotesize{0.418~\scriptsize{$\pm$.00}} & {\footnotesize{\bf 0.159}~\scriptsize{$\pm$.00}} &  
    \footnotesize{33.39~\scriptsize{$\pm$2.8}} & {\footnotesize{\bf 0.459}~\scriptsize{$\pm$.19}}  \\

    \midrule
    &$n$& 
    \multicolumn{8}{c}{\it IHDP ($d=25$)  with increasing sample size}\\
    \midrule
    
    &$100$ &             
    \footnotesize{1.608~\scriptsize{$\pm$.02}} & {\footnotesize{\bf 1.565}~\scriptsize{$\pm$.25}} & 
    \footnotesize{8.076~\scriptsize{$\pm$2.4}} & {\footnotesize{\bf 1.956}~\scriptsize{$\pm$.64}} & 
    \footnotesize{1.944~\scriptsize{$\pm$.00}} & {\cellcolor{green!10}\footnotesize{\bf 1.542}~\scriptsize{$\pm$.18}} &  
    \footnotesize{24.53~\scriptsize{$\pm$5.1}} & {\footnotesize{\bf 11.52}~\scriptsize{$\pm$8.4}}  \\
    
    &$250$ &
    \footnotesize{2.335~\scriptsize{$\pm$.00}} & {\footnotesize{\bf 1.637}~\scriptsize{$\pm$.02}} & 
    \footnotesize{8.607~\scriptsize{$\pm$.82}} & {\footnotesize{\bf 1.964}~\scriptsize{$\pm$.49}} & 
    \footnotesize{2.219~\scriptsize{$\pm$.00}} & {\cellcolor{green!10}\footnotesize{\bf 1.627}~\scriptsize{$\pm$.03}} &  
    \footnotesize{35.37~\scriptsize{$\pm$3.8}} & {\footnotesize{\bf 18.13}~\scriptsize{$\pm$7.8}}  \\
    
    &$500$&
    \footnotesize{2.177~\scriptsize{$\pm$.00}} & {\footnotesize{\bf 1.536}~\scriptsize{$\pm$.00}} & 
    \footnotesize{4.216~\scriptsize{$\pm$.35}} & {\footnotesize{\bf 1.739}~\scriptsize{$\pm$.36}} & 
    \footnotesize{2.233~\scriptsize{$\pm$.00}} & {\cellcolor{green!10}\footnotesize{\bf 1.535}~\scriptsize{$\pm$.00}} &  
    \footnotesize{26.06~\scriptsize{$\pm$4.5}} & {\footnotesize{\bf 10.18}~\scriptsize{$\pm$5.6}}  \\
    
    \bottomrule
    \end{tabularx}
    \end{center}
\end{table*}

\renewcommand{\arraystretch}{0.5}
\setlength{\tabcolsep}{1.5pt}
\begin{table*}[t]
    \caption{{\bf Results on (semi-)synthetic data (Twins \& IHDP) with \texttt{Ridge Regression}.} We report for the same configuration as in \Cref{tab:res:CATE:synth}. Results are averaged over ten runs with (\quotes{\cmark}), and without (\quotes{\xmark}) the same representations used in \Cref{tab:res:CATE:synth}.}
    \label{tab:res:CATE:extra:lin}
        \vspace{-13pt}
\begin{center}
    \begin{tabularx}{\textwidth}{l r @{\hskip 2pt}  *{4}{|CC}}
    
    \toprule
    \multicolumn{2}{r|}{Methods}   
        & \multicolumn{2}{c}{\bf X-Learner}  
        & \multicolumn{2}{c}{\bf DR-Learner} 
        & \multicolumn{2}{c}{\bf T-Learner} 
        & \multicolumn{2}{c}{\bf R-Learner}\\
    \midrule     
    \multicolumn{2}{r|}{EBM} & \xmark & \cmark & \xmark & \cmark & \xmark & \cmark & \xmark & \cmark \\
    
    \midrule  
    
      $d$ & $n$& 
    \multicolumn{8}{c}{\it Synth. data with increasing sample size and increasing dimensions}\\
     \midrule   
    
    $50$ &$100$ & 
    \footnotesize{2.373~\scriptsize{$\pm$.00}} & {\footnotesize{\bf 2.001}~\scriptsize{$\pm$.02}} & 
    \footnotesize{10.53~\scriptsize{$\pm$4.8}} & {\footnotesize{\bf 2.028}~\scriptsize{$\pm$.06}} & 
    \footnotesize{2.471~\scriptsize{$\pm$.00}} & {\cellcolor{green!10}\footnotesize{\bf 1.997}~\scriptsize{$\pm$.02}} &  
    \footnotesize{53.24~\scriptsize{$\pm$5.3}} & {\footnotesize{\bf 11.33}~\scriptsize{$\pm$3.9}}  \\
    
    $100$ &$250$ & 
    \footnotesize{2.802~\scriptsize{$\pm$.00}} & {\cellcolor{green!10}\footnotesize{\bf 2.021}~\scriptsize{$\pm$.01}} & 
    \footnotesize{8.769~\scriptsize{$\pm$1.5}} & {\footnotesize{\bf 2.041}~\scriptsize{$\pm$.05}} & 
    \footnotesize{2.871~\scriptsize{$\pm$.00}} & {\footnotesize{\bf 2.021}~\scriptsize{$\pm$.01}} &  
    \footnotesize{130.6~\scriptsize{$\pm$75.}} & {\footnotesize{\bf 22.30}~\scriptsize{$\pm$8.3}}  \\
    
     $150$ &$500$ &
    \footnotesize{2.581~\scriptsize{$\pm$.00}} & {\cellcolor{green!10}\footnotesize{\bf 2.001}~\scriptsize{$\pm$.01}} & 
    \footnotesize{3.074~\scriptsize{$\pm$.10}} & {\footnotesize{\bf 2.001}~\scriptsize{$\pm$.01}} & 
    \footnotesize{2.601~\scriptsize{$\pm$.00}} & {\footnotesize{\bf 2.001}~\scriptsize{$\pm$.01}} &  
    \footnotesize{50.13~\scriptsize{$\pm$4.1}} & {\footnotesize{\bf 12.31}~\scriptsize{$\pm$1.8}}  \\
    
    $200$ &    $1$k & 
    \footnotesize{2.187~\scriptsize{$\pm$.00}} & {\footnotesize{\bf 1.942}~\scriptsize{$\pm$.01}} & 
    \footnotesize{2.304~\scriptsize{$\pm$.03}} & {\cellcolor{green!10}\footnotesize{\bf 1.941}~\scriptsize{$\pm$.01}} & 
    \footnotesize{2.192~\scriptsize{$\pm$.00}} & {\footnotesize{\bf 1.941}~\scriptsize{$\pm$.01}} &  
    \footnotesize{42.77~\scriptsize{$\pm$3.8}} & {\footnotesize{\bf 2.839}~\scriptsize{$\pm$.62}}  \\
    
     $250$ &$1.5$k &
    \footnotesize{2.270~\scriptsize{$\pm$.00}} & {\footnotesize{\bf 1.958}~\scriptsize{$\pm$.02}} & 
    \footnotesize{2.412~\scriptsize{$\pm$.04}} & {\footnotesize{\bf 1.977}~\scriptsize{$\pm$.03}} & 
    \footnotesize{2.274~\scriptsize{$\pm$.00}} & {\cellcolor{green!10}\footnotesize{\bf 1.957}~\scriptsize{$\pm$.02}} &  
    \footnotesize{42.03~\scriptsize{$\pm$3.5}} & {\footnotesize{\bf 2.511}~\scriptsize{$\pm$.17}}  \\
    
    \midrule
    &$n$& 
    \multicolumn{8}{c}{\it Synth. data with increasing sample size and dimensions fixed at $d=100$}\\
    \midrule
    
     &$100$& 
    \footnotesize{2.124~\scriptsize{$\pm$.00}} & {\cellcolor{green!10}\footnotesize{\bf 1.946}~\scriptsize{$\pm$.03}} & 
    \footnotesize{78.12~\scriptsize{$\pm$24.}} & {\footnotesize{\bf 2.145}~\scriptsize{$\pm$.11}} & 
    \footnotesize{2.272~\scriptsize{$\pm$.00}} & {\footnotesize{\bf 1.948}~\scriptsize{$\pm$.04}} &  
    \footnotesize{58.24~\scriptsize{$\pm$4.9}} & {\footnotesize{\bf 5.346}~\scriptsize{$\pm$1.9}}  \\
    
    &$250$ & 
    \footnotesize{2.802~\scriptsize{$\pm$.00}} & {\cellcolor{green!10}\footnotesize{\bf 2.018}~\scriptsize{$\pm$.00}} & 
    \footnotesize{4.489~\scriptsize{$\pm$.26}} & {\footnotesize{\bf 2.025}~\scriptsize{$\pm$.01}} & 
    \footnotesize{2.871~\scriptsize{$\pm$.00}} & {\footnotesize{\bf 2.019}~\scriptsize{$\pm$.01}} &  
    \footnotesize{60.27~\scriptsize{$\pm$4.1}} & {\footnotesize{\bf 9.607}~\scriptsize{$\pm$3.0}}  \\
     
    &$500$ &
    \footnotesize{2.152~\scriptsize{$\pm$.00}} & {\footnotesize{\bf 2.059}~\scriptsize{$\pm$.04}} & 
    \footnotesize{2.373~\scriptsize{$\pm$.08}} & {\cellcolor{green!10}\footnotesize{\bf 2.003}~\scriptsize{$\pm$.05}} & 
    \footnotesize{2.164~\scriptsize{$\pm$.00}} & {\footnotesize{\bf 2.056}~\scriptsize{$\pm$.04}} &  
    \footnotesize{42.13~\scriptsize{$\pm$3.5}} & {\footnotesize{\bf 3.755}~\scriptsize{$\pm$.83}}  \\

    &$1$k  &
    \footnotesize{2.062~\scriptsize{$\pm$.00}} & {\footnotesize{\bf 1.984}~\scriptsize{$\pm$.02}} & 
    \footnotesize{2.109~\scriptsize{$\pm$.01}} & {\footnotesize{\bf 2.022}~\scriptsize{$\pm$.09}} & 
    \footnotesize{2.064~\scriptsize{$\pm$.00}} & {\cellcolor{green!10}\footnotesize{\bf 1.983}~\scriptsize{$\pm$.02}} &  
    \footnotesize{39.49~\scriptsize{$\pm$3.2}} & {\footnotesize{\bf 2.608}~\scriptsize{$\pm$.19}}  \\

    &$1.5$k &
    \footnotesize{2.013~\scriptsize{$\pm$.00}} & {\footnotesize{\bf 2.003}~\scriptsize{$\pm$.02}} & 
    \footnotesize{2.052~\scriptsize{$\pm$.01}} & {\footnotesize{\bf 2.398}~\scriptsize{$\pm$1.2}} & 
    \footnotesize{2.014~\scriptsize{$\pm$.00}} & {\cellcolor{green!10}\footnotesize{\bf 2.001}~\scriptsize{$\pm$.02}} &  
    \footnotesize{37.19~\scriptsize{$\pm$2.9}} & {\footnotesize{\bf 3.954}~\scriptsize{$\pm$.77}}  \\

    \midrule
    &$n$& 
    \multicolumn{8}{c}{\it Twins ($d=48$)  with increasing sample size}\\
    \midrule
    
    &$500$ &             
    \footnotesize{0.182~\scriptsize{$\pm$.00}} & {\cellcolor{green!10}\footnotesize{\bf 0.151}~\scriptsize{$\pm$.00}} & 
    \footnotesize{1.161~\scriptsize{$\pm$.20}} & {\footnotesize{\bf 0.168}~\scriptsize{$\pm$.02}} & 
    \footnotesize{0.183~\scriptsize{$\pm$.00}} & {\footnotesize{\bf 0.151}~\scriptsize{$\pm$.00}} &  
    \footnotesize{134.6~\scriptsize{$\pm$12.}} & {\footnotesize{\bf 0.842}~\scriptsize{$\pm$.22}}  \\
    
    &$1$k &
    \footnotesize{0.196~\scriptsize{$\pm$.00}} & {\cellcolor{green!10}\footnotesize{\bf 0.159}~\scriptsize{$\pm$.00}} & 
    \footnotesize{0.261~\scriptsize{$\pm$.02}} & {\footnotesize{\bf 0.172}~\scriptsize{$\pm$.01}} & 
    \footnotesize{0.196~\scriptsize{$\pm$.00}} & {\footnotesize{\bf 0.159}~\scriptsize{$\pm$.00}} &  
    \footnotesize{58.70~\scriptsize{$\pm$4.4}} & {\footnotesize{\bf 0.455}~\scriptsize{$\pm$.14}}  \\
    
    &$1.5$k&
    \footnotesize{0.166~\scriptsize{$\pm$.00}} & {\cellcolor{green!10}\footnotesize{\bf 0.156}~\scriptsize{$\pm$.00}} & 
    \footnotesize{0.171~\scriptsize{$\pm$.01}} & {\footnotesize{\bf 0.159}~\scriptsize{$\pm$.00}} & 
    \footnotesize{0.166~\scriptsize{$\pm$.00}} & {\footnotesize{\bf 0.156}~\scriptsize{$\pm$.00}} &  
    \footnotesize{29.72~\scriptsize{$\pm$2.4}} & {\footnotesize{\bf 0.415}~\scriptsize{$\pm$.14}}  \\

    &$2$k   &
    \footnotesize{0.163~\scriptsize{$\pm$.00}} & {\cellcolor{green!10}\footnotesize{\bf 0.153}~\scriptsize{$\pm$.00}} & 
    \footnotesize{0.319~\scriptsize{$\pm$.04}} & {\footnotesize{\bf 0.157}~\scriptsize{$\pm$.00}} & 
    \footnotesize{0.163~\scriptsize{$\pm$.00}} & {\footnotesize{\bf 0.153}~\scriptsize{$\pm$.00}} &  
    \footnotesize{176.0~\scriptsize{$\pm$13.}} & {\footnotesize{\bf 0.601}~\scriptsize{$\pm$.30}}  \\
    
    &$2.5$k &
    \footnotesize{0.169~\scriptsize{$\pm$.00}} & {\cellcolor{green!10}\footnotesize{\bf 0.162}~\scriptsize{$\pm$.00}} & 
    \footnotesize{0.321~\scriptsize{$\pm$.04}} & {\footnotesize{\bf 0.164}~\scriptsize{$\pm$.00}} & 
    \footnotesize{0.169~\scriptsize{$\pm$.00}} & {\footnotesize{\bf 0.162}~\scriptsize{$\pm$.00}} &  
    \footnotesize{207.8~\scriptsize{$\pm$17.}} & {\footnotesize{\bf 0.479}~\scriptsize{$\pm$.19}}  \\
    
    \midrule
    &$n$& 
    \multicolumn{8}{c}{\it IHDP ($d=25$)  with increasing sample size}\\
    \midrule
    
    &$100$ &             
    \footnotesize{1.807~\scriptsize{$\pm$.00}} & {\cellcolor{green!10}\footnotesize{\bf 1.673}~\scriptsize{$\pm$.02}} & 
    \footnotesize{5.933~\scriptsize{$\pm$1.2}} & {\footnotesize{\bf 2.456}~\scriptsize{$\pm$.47}} & 
    \footnotesize{1.739~\scriptsize{$\pm$.00}} & {\footnotesize{\bf 1.675}~\scriptsize{$\pm$.02}} &  
    \footnotesize{23.56~\scriptsize{$\pm$3.2}} & {\footnotesize{\bf 14.19}~\scriptsize{$\pm$9.3}}  \\
    
    &$250$ &
    \footnotesize{1.659~\scriptsize{$\pm$.00}} & {\footnotesize{\bf 1.579}~\scriptsize{$\pm$.03}} & 
    \footnotesize{2.883~\scriptsize{$\pm$.12}} & {\footnotesize{\bf 2.426}~\scriptsize{$\pm$.09}} & 
    \footnotesize{1.693~\scriptsize{$\pm$.00}} & {\cellcolor{green!10}\footnotesize{\bf 1.577}~\scriptsize{$\pm$.03}} &  
    \footnotesize{11.01~\scriptsize{$\pm$2.0}} & {\footnotesize{\bf 7.985}~\scriptsize{$\pm$3.1}}  \\
    
    &$500$&
    \footnotesize{1.625~\scriptsize{$\pm$.00}} & {\footnotesize{\bf 1.614}~\scriptsize{$\pm$.01}} & 
    \footnotesize{1.819~\scriptsize{$\pm$.16}} & {\footnotesize{\bf 1.673}~\scriptsize{$\pm$.09}} & 
    \footnotesize{1.641~\scriptsize{$\pm$.00}} & {\cellcolor{green!10}\footnotesize{\bf 1.610}~\scriptsize{$\pm$.02}} &  
    \footnotesize{6.614~\scriptsize{$\pm$.59}} & {\footnotesize{\bf 5.602}~\scriptsize{$\pm$1.6}}  \\
    
    \bottomrule
    \end{tabularx}
    \end{center}
\end{table*}

\renewcommand{\arraystretch}{0.5}
\setlength{\tabcolsep}{1.5pt}
\begin{table*}[t]
    \caption{{\bf Copy of \Cref{tab:res:CATE:synth} with additional data (IHDP).} To save space, we include the \quotes{complete} table of our main experiment here, in our supplemental material. The content of this table is exactly the same as in \Cref{tab:res:CATE:synth}, except for added results on IHDP (bottom block).}
       \vspace{-13pt}
    \label{tab:res:CATE:ihdp_main}
    \begin{center}
    \begin{tabularx}{\textwidth}{l r @{\hskip 2pt}  *{4}{|CC}}
    
   \midrule
   \multicolumn{2}{r|}{Methods}   & \multicolumn{2}{c}{\bf X-Learner}  
      & \multicolumn{2}{c}{\bf DR-Learner} 
      & \multicolumn{2}{c}{\bf T-Learner} 
      & \multicolumn{2}{c}{\bf R-Learner}\\
  \midrule     
 \multicolumn{2}{r|}{EBM} & \xmark & \cmark & \xmark & \cmark & \xmark & \cmark & \xmark & \cmark \\

    \toprule

      $d$ & $n$& 
    \multicolumn{8}{c}{\it Synth. data with increasing sample size and increasing dimensions}\\
     \midrule   
    
    $50$ &$100$ & 
    \footnotesize{2.309~\scriptsize{$\pm$.00}} & {\footnotesize{\bf 1.994}~\scriptsize{$\pm$.02}} & 
    \footnotesize{4.594~\scriptsize{$\pm$.56}} & {\footnotesize{\bf 2.017}~\scriptsize{$\pm$.04}} & 
    \footnotesize{2.441~\scriptsize{$\pm$.00}} & {\footnotesize{\bf 1.993}~\scriptsize{$\pm$.01}} & 
    \footnotesize{3.194~\scriptsize{$\pm$.26}} & {\cellcolor{green!10} \footnotesize{\bf 1.982}~\scriptsize{$\pm$.04}} \\
    
    $100$ &$250$ & 
    \footnotesize{2.779~\scriptsize{$\pm$.00}} & {\footnotesize{\bf 2.018}~\scriptsize{$\pm$.01}} &  
    \footnotesize{4.056~\scriptsize{$\pm$.32}} & {\footnotesize{\bf 2.154}~\scriptsize{$\pm$.39}} & 
    \footnotesize{2.838~\scriptsize{$\pm$.00}} & {\footnotesize{\bf 2.019}~\scriptsize{$\pm$.01}} & 
    \footnotesize{3.702~\scriptsize{$\pm$.23}} & {\cellcolor{green!10}\footnotesize{\bf 2.018}~\scriptsize{$\pm$.01}} \\
    
     $150$ &$500$ &
    \footnotesize{2.618~\scriptsize{$\pm$.00}} & {\footnotesize{\bf 2.000}~\scriptsize{$\pm$.01}} & 
    \footnotesize{3.030~\scriptsize{$\pm$.12}} & {\footnotesize{\bf 2.001}~\scriptsize{$\pm$.01}} & 
    \footnotesize{2.641~\scriptsize{$\pm$.00}} & {\cellcolor{green!10}\footnotesize{\bf 2.000}~\scriptsize{$\pm$.01}} & 
    \footnotesize{2.877~\scriptsize{$\pm$.08}} & {\footnotesize{\bf 2.000}~\scriptsize{$\pm$.01}} \\
    
    $200$ &    $1$k & 
    \footnotesize{2.185~\scriptsize{$\pm$.00}} & {\footnotesize{\bf 1.940}~\scriptsize{$\pm$.01}} & 
    \footnotesize{2.283~\scriptsize{$\pm$.02}} & {\footnotesize{\bf 1.941}~\scriptsize{$\pm$.01}} & 
    \footnotesize{2.189~\scriptsize{$\pm$.00}} & {\cellcolor{green!10}\footnotesize{\bf 1.939}~\scriptsize{$\pm$.01}} & 
    \footnotesize{2.271~\scriptsize{$\pm$.01}} & {\footnotesize{\bf 1.940}~\scriptsize{$\pm$.01}} \\
    
     $250$ &$1.5$k &
    \footnotesize{2.267~\scriptsize{$\pm$.00}} & {\footnotesize{\bf 1.949}~\scriptsize{$\pm$.02}} & 
    \footnotesize{2.427~\scriptsize{$\pm$.01}} & {\footnotesize{\bf 1.976}~\scriptsize{$\pm$.00}} & 
    \footnotesize{2.271~\scriptsize{$\pm$.00}} & {\cellcolor{green!10}\footnotesize{\bf 1.948}~\scriptsize{$\pm$.01}} & 
    \footnotesize{2.436~\scriptsize{$\pm$.02}} & {\footnotesize{\bf 1.949}~\scriptsize{$\pm$.02}} \\
    
    \midrule
    &$n$& 
    \multicolumn{8}{c}{\it Synth. data with increasing sample size and dimensions fixed at $d=100$}\\
    \midrule
    
     &$100$& 
    \footnotesize{2.134~\scriptsize{$\pm$.00}} & {\footnotesize{\bf 1.927}~\scriptsize{$\pm$.01}} & 
    \footnotesize{24.61~\scriptsize{$\pm$9.9}} & {\footnotesize{\bf 2.096}~\scriptsize{$\pm$.09}} & 
    \footnotesize{2.279~\scriptsize{$\pm$.00}} & {\footnotesize{\bf 1.929}~\scriptsize{$\pm$.01}} & 
    \footnotesize{3.192~\scriptsize{$\pm$.13}} & {\cellcolor{green!10}\footnotesize{\bf 1.925}~\scriptsize{$\pm$.01}} \\
    
     &$250$&
    \footnotesize{2.779~\scriptsize{$\pm$.00}} & {\footnotesize{\bf 2.018}~\scriptsize{$\pm$.01}} &  
    \footnotesize{4.056~\scriptsize{$\pm$.32}} & {\footnotesize{\bf 2.154}~\scriptsize{$\pm$.39}} & 
    \footnotesize{2.838~\scriptsize{$\pm$.00}} & {\footnotesize{\bf 2.019}~\scriptsize{$\pm$.01}} & 
    \footnotesize{3.702~\scriptsize{$\pm$.23}} & {\cellcolor{green!10}\footnotesize{\bf 2.018}~\scriptsize{$\pm$.01}} \\
     
    &$500$ &
    \footnotesize{2.155~\scriptsize{$\pm$.00}} & {\footnotesize{\bf 2.056}~\scriptsize{$\pm$.02}} & 
    \footnotesize{2.334~\scriptsize{$\pm$.07}} & {\footnotesize{\bf 2.273}~\scriptsize{$\pm$.67}} & 
    \footnotesize{2.166~\scriptsize{$\pm$.00}} & {\cellcolor{green!10}\footnotesize{\bf 2.053}~\scriptsize{$\pm$.02}} & 
    \footnotesize{2.271~\scriptsize{$\pm$.05}} & {\footnotesize{\bf 2.056}~\scriptsize{$\pm$.02}} \\

    &$1$k  &
    \footnotesize{2.059~\scriptsize{$\pm$.00}} & {\cellcolor{green!10}\footnotesize{\bf 1.964}~\scriptsize{$\pm$.02}} &
    \footnotesize{2.105~\scriptsize{$\pm$.01}} & {\footnotesize{\bf 2.016}~\scriptsize{$\pm$.16}} & 
    \footnotesize{2.061~\scriptsize{$\pm$.00}} & {\footnotesize{\bf 1.964}~\scriptsize{$\pm$.02}} & 
    \footnotesize{2.086~\scriptsize{$\pm$.01}} & {\footnotesize{\bf 1.965}~\scriptsize{$\pm$.02}} \\

    &$1.5$k &
    \footnotesize{2.013~\scriptsize{$\pm$.00}} & {\footnotesize{\bf 1.998}~\scriptsize{$\pm$.02}} & 
    \footnotesize{2.043~\scriptsize{$\pm$.01}} & {\footnotesize{\bf 1.998}~\scriptsize{$\pm$.02}} & 
    \footnotesize{2.014~\scriptsize{$\pm$.00}} & {\footnotesize{\bf 1.998}~\scriptsize{$\pm$.02}} & 
    \footnotesize{2.024~\scriptsize{$\pm$.01}} & {\cellcolor{green!10}\footnotesize{\bf 1.991}~\scriptsize{$\pm$.02}} \\

    \midrule
    &$n$& 
    \multicolumn{8}{c}{\it Twins ($d=48$)  with increasing sample size}\\
    \midrule
    
    &$500$ &             
    \footnotesize{0.214~\scriptsize{$\pm$.00}} & {\cellcolor{green!10}\footnotesize{\bf 0.144}~\scriptsize{$\pm$.00}} & 
    \footnotesize{0.236~\scriptsize{$\pm$.04}} & {\footnotesize{\bf 0.182}~\scriptsize{$\pm$.05}} & 
    \footnotesize{0.221~\scriptsize{$\pm$.00}} & {\footnotesize{\bf 0.145}~\scriptsize{$\pm$.00}} & 
    \footnotesize{0.222~\scriptsize{$\pm$.02}} & {\footnotesize{\bf 0.145}~\scriptsize{$\pm$.00}} \\
    
    &$1$k &
    \footnotesize{0.294~\scriptsize{$\pm$.00}} & {\footnotesize{\bf 0.162}~\scriptsize{$\pm$.00}} & 
    \footnotesize{0.348~\scriptsize{$\pm$.12}} & {\footnotesize{\bf 0.173}~\scriptsize{$\pm$.03}} & 
    \footnotesize{0.301~\scriptsize{$\pm$.00}} & {\footnotesize{\bf 0.162}~\scriptsize{$\pm$.01}} & 
    \footnotesize{0.532~\scriptsize{$\pm$.11}} & {\footnotesize{\cellcolor{green!10}\bf 0.161}~\scriptsize{$\pm$.00}} \\
    
    &$1.5$k&
    \footnotesize{0.165~\scriptsize{$\pm$.00}} & {\footnotesize{\bf 0.154}~\scriptsize{$\pm$.00}} & 
    \footnotesize{0.189~\scriptsize{$\pm$.06}} & {\footnotesize{\bf  0.159}~\scriptsize{$\pm$.01}} & 
    \footnotesize{0.165~\scriptsize{$\pm$.00}} & {\footnotesize{\bf 0.154}~\scriptsize{$\pm$.00}} & 
    \footnotesize{0.172~\scriptsize{$\pm$.01}} & {\footnotesize{\cellcolor{green!10}\bf 0.154}~\scriptsize{$\pm$.00}} \\

    &$2$k   &
    \footnotesize{0.167~\scriptsize{$\pm$.00}} & {\cellcolor{green!10}\footnotesize{\bf 0.156}~\scriptsize{$\pm$.00}} & 
    \footnotesize{0.197~\scriptsize{$\pm$.03}} & {\footnotesize{\bf 0.159}~\scriptsize{$\pm$.00}} & 
    \footnotesize{0.167~\scriptsize{$\pm$.00}} & {\footnotesize{\bf 0.156}~\scriptsize{$\pm$.00}} & 
    \footnotesize{0.222~\scriptsize{$\pm$.05}} & {\footnotesize{\bf 0.157}~\scriptsize{$\pm$.00}} \\
    
    &$2.5$k &
    \footnotesize{0.297~\scriptsize{$\pm$.00}} & {\footnotesize{\bf 0.153}~\scriptsize{$\pm$.00}} & 
    \footnotesize{0.390~\scriptsize{$\pm$.19}} & {\footnotesize{\bf 0.156}~\scriptsize{$\pm$.00}} & 
    \footnotesize{0.297~\scriptsize{$\pm$.00}} & {\footnotesize{\bf 0.153}~\scriptsize{$\pm$.00}} & 
    \footnotesize{0.358~\scriptsize{$\pm$.22}} & {\cellcolor{green!10}\footnotesize{\bf 0.153}~\scriptsize{$\pm$.00}} \\
    
    \midrule
    &$n$& 
    \multicolumn{8}{c}{\it IHDP ($d=25$)  with increasing sample size}\\
    \midrule
    
    &$100$ &             
    \footnotesize{3.369~\scriptsize{$\pm$.36}} & {\cellcolor{green!10}\footnotesize{\bf 1.783}~\scriptsize{$\pm$.22}} & 
    \footnotesize{6.484~\scriptsize{$\pm$5.7}} & {\footnotesize{\bf 2.329}~\scriptsize{$\pm$.67}} & 
    \footnotesize{6.138~\scriptsize{$\pm$.50}} & {\footnotesize{\bf 2.028}~\scriptsize{$\pm$.65}} &  
    \footnotesize{22.57~\scriptsize{$\pm$4.3}} & {\footnotesize{\bf 16.43}~\scriptsize{$\pm$9.1}}  \\

    &$250$ &
    \footnotesize{47.29~\scriptsize{$\pm$.64}} & {\footnotesize{\bf 2.788}~\scriptsize{$\pm$1.2}} & 
    \footnotesize{1.899~\scriptsize{$\pm$.10}} & {\cellcolor{green!10}\footnotesize{\bf 1.689}~\scriptsize{$\pm$.26}} & 
    \footnotesize{51.76~\scriptsize{$\pm$.64}} & {\footnotesize{\bf 2.963}~\scriptsize{$\pm$1.3}} &  
    \footnotesize{72.20~\scriptsize{$\pm$9.4}} & {\footnotesize{\bf 54.16}~\scriptsize{$\pm$22.}}  \\
    
    &$500$&
    \footnotesize{2.176~\scriptsize{$\pm$.25}} & {\cellcolor{green!10}\footnotesize{\bf 1.532}~\scriptsize{$\pm$.01}} & 
    \footnotesize{1.681~\scriptsize{$\pm$.04}} & {\footnotesize{\bf 1.554}~\scriptsize{$\pm$.02}} & 
    \footnotesize{4.361~\scriptsize{$\pm$.35}} & {\footnotesize{\bf 1.552}~\scriptsize{$\pm$.03}} &  
    \footnotesize{8.531~\scriptsize{$\pm$.94}} & {\footnotesize{\bf 4.661}~\scriptsize{$\pm$.75}}  \\
    \bottomrule
    \end{tabularx}
    \end{center}
\end{table*}

\clearpage
\newpage

\renewcommand{\arraystretch}{0.4}
\setlength{\tabcolsep}{1.5pt}
\begin{table*}[t]
    \centering
    \caption{{\bf Chosen hyperparameters for \Cref{tab:res:CATE:synth}.} We performed hyperparamter sweeps for each setup using a Bayesian optimisation scheme \citep{wandb}. Our searched ranges are reported in \Cref{tab:hyperparams:sweep_ranges}. We have rounded continuous hyperparameters to user-friendly values (as they are sampled from continuous distributions during optimization). Twins settings were also used in \Cref{fig:three graphs}, but with a fixed $k=5$ for both AE and EBM. Each integer (separated by a dash) in \quotes{Architecture} indicates layer width; \quotes{20-20} thus means a neural network with two hidden layers, each of width 20.}
     \vspace{-9pt}
    \label{tab:hyperparams}
    \begin{tabularx}{\textwidth}{lr | *{4}{C}}
    \toprule
    \multicolumn{2}{l}{Setup} & $b$ & $k$ & Architecture & Perturbation prob. \\
    \toprule
    $d$ & $n$ & \multicolumn{4}{c}{Synth. data, increasing dim}\\
    \midrule
    50 & 100 & 10 & 3 & 20-20-20 & 0.20 \\
    100 & 250 & 10 & 4 & 20-20-20 & 0.50 \\
    150& 500 & 5 & 3 & 20-20 & 0.20 \\
    200 & 1k & 3 & 15 & 20-20-20-20 & 0.50 \\
    250 & 1.5k & 3 & 20 & 20-20-20 & 0.50\\
    \midrule
    & $n$ & \multicolumn{4}{c}{Synth. data, fixed dim ($d$=100)}\\
    \midrule
    & 100 & 5 & 15 & 20-20-20-20-20-20 & 0.20 \\
    & 250 & 10 & 4 & 20-20-20 & 0.50 \\
    & 500 & 3 & 10 & 20-20-20-20 & 0.50 \\
    & 1k & 3 & 20 & 20-20 & 0.35 \\
    & 1.5k & 3 & 10 & 20-20 & 0.30 \\
    \midrule
    & $n$ & \multicolumn{4}{c}{Twins, increasing $n$} \\
    \midrule
    & 500 & 5 & 15 & 20-20-20-20-20-20 & 0.45 \\
    & 1k & 5 & 16 & 20-20-20-20-20-20 & 0.55 \\
    & 1.5k & 5 & 16 & 20-20-20-20-20-20 & 0.55 \\
    & 2k & 4 & 14 & 20-20-20-20-20-20 & 0.55 \\
    & 2.5k & 4 & 12 & 20-20-20-20-20-20 & 0.50 \\
    
    \midrule
    & $n$ & \multicolumn{4}{c}{IHDP, increasing $n$} \\
    \midrule
    & 100 & 1 & 5 & 36-36-36-36-36-36 & 0.45 \\
    & 250 & 1 & 5 & 36-36-36-36-36-36 & 0.45 \\
    & 500 & 1 & 5 & 36-36-36-36-36-36 & 0.45 \\

    \bottomrule
    \end{tabularx}
\end{table*}

\renewcommand{\arraystretch}{0.4}
\setlength{\tabcolsep}{1.5pt}
\begin{table*}[t]
    \centering
    \caption{{\bf Ranges for hyperparameter sweeps, for \Cref{tab:hyperparams}.} For each setup: (I) Synth. data, increasing dim, (II) Synth. data, fixed dim ($d$=100) (III) Twins, increasing dim, and (IV) IHDP, increasing dim;   we used a Bayesian optimization (BO) scheme to find our selected hyperparameters. We chose BO as training representations can get expensive. In or BO setup, we maximized the loss \eqref{equ:rank} on a (20\%) validation-set. }
    \label{tab:hyperparams:sweep_ranges}
    \vspace{-9pt}
    \begin{tabularx}{\textwidth}{l | *{4}{C}}
    \toprule
    Setup & $b$ & $k$ & \# layers & Perturbation prob. \\
    \toprule
    (I) &$\mathcal{U}(1;2;...;10)$ &$\mathcal{U}(3;4;...;25)$&$\mathcal{U}(2;3;4;5;6)$& $\mathcal{U}(0.2; 0.8)$ \\
    (II) &$\mathcal{U}(1;2;...;10)$ &$\mathcal{U}(3;4;...;25)$&$\mathcal{U}(2;3;4;5;6)$& $\mathcal{U}(0.2; 0.8)$ \\
    (III) &$\mathcal{U}(1;2;...;10)$ &$\mathcal{U}(3;4;...;25)$&$\mathcal{U}(2;3;4;5;6)$& $\mathcal{U}(0.2; 0.8)$ \\
    (IV) &$\mathcal{U}(1;2;...;10)$ &$\mathcal{U}(3;4;...;25)$&$\mathcal{U}(2;3;4;5;6)$& $\mathcal{U}(0.2; 0.8)$ \\
    \bottomrule
    \end{tabularx}
\end{table*}

\subsection{Hyperparameters} \label{sect:app:hyper_details}
We report our chosen hyperparameters for each sample-size in \Cref{tab:hyperparams}. We found these values through a Bayesian optimization scheme \citep{wandb}. The used ranges are reported in \Cref{tab:hyperparams:sweep_ranges}. As an insight, we noticed that the architecture and amount of noisy samples made little difference to performance in PEHE. The perturbation probability, and the value of $k$ {\it did} make a difference, especially in larger sample sizes.  Each experiment was performed on an Nvidia \texttt{GeForce RTX 2080 Ti} GPU, and 6 Intel \texttt{i5-8600K (3.60GHz)} CPUs.  In some instances we ran independent experiments on a duplicate system.

\subsection{Details on benchmarked CATE estimators} \label{sect:app:cate_details}
We use EconML \citep{econml}\footnote{EconML is available open-source under an MIT License. Please find all details on their repository, \url{https://github.com/microsoft/EconML}.} to evaluate the CATE learners. We keep the hyperparameters for each CATE learner as their default, except for the  regression models (which by default are linear). In \Cref{tab:res:CATE:synth} we replace each regressor by a \texttt{KernelRidge} regressor, and each classifier by a support vector machine (\texttt{SVC}); both implemented by \citet{sklearn}. Results on CATE learners with alternative models are in \Cref{sect:additonal:ML}.

\clearpage
\newpage

\section{Proofs}\label{sect:proof}

\subsection{Proof of \Cref{prop:suff}}
\begin{proof}
The sufficiency of $f_{\theta}$ is obtained immediately by applying Fisher–Neyman factorization theorem to \eqref{equ:model}. Consider another sufficient statistics $\tilde{u}:\mathcal{X}\rightarrow\mathcal{U}$. By the factorization theorem, we can factorize the distribution as $p_{\theta,j}(x) = \tilde{h}(x)\tilde{g}(\tilde{u}(x),\beta_j)$ for some functions $\tilde{h}$ and $\tilde{g}$. Given any $x$ and $x'$ such that $\tilde{u}(x) = \tilde{u}(x')$, i.e., $\tilde{g}(\tilde{u}(x),\beta_j) = \tilde{g}(\tilde{u}(x'),\beta_j)$, we have
$\exp\left[-\beta_j^{\top}(f_{\theta}(x) - f_{\theta}(x'))\right]=\frac{p_{\theta,j}(x)}{p_{\theta,j}(x')} =  \frac{\tilde{h}(x)}{\tilde{h}(x')}.$ 

The first equality uses two facts of the standard EBM: (1) $h(x)$ is constant ($h(x) = 1$), and (2) $Z_{\theta,j}$ is for fixed $\beta_j$. The ratio on the right-hand side does not depend on $\beta_j$, which holds if and only if $f_{\theta}(x) = f_{\theta}(x')$. That is, for any sufficient statistics $\tilde{u}$ and any $x,x'$, $\tilde{u}(x) = \tilde{u}(x')$ implies that $f_{\theta}(x) = f_{\theta}(x')$ and hence that $f_{\theta}(x)$ is a function of $\tilde{u}(x)$. Then $f_{\theta}$ is a minimal sufficient statistic.
\end{proof}

\subsection{Proof of \Cref{prop:id}} \label{sect:app:proof33}

\begin{proof}
Consider two different parameter values $\theta $ and $\tilde{\theta}$ such that
$p_{\theta,j}(x) = p_{\tilde{\theta},j}(x)$.
Using the expression \eqref{equ:model} and applying logarithm to both sides,
\begin{equation}\label{equ:proof}
\beta_j^{\top} f_{\theta}(x) = \beta_j^{\top} f_{\tilde{\theta}}(x) + \log \tfrac{Z_{ \tilde{\theta},j}}{Z_{\theta,j}}. 
\end{equation}
By concatenating the last equation for all $j\in [k]$, we have
\begin{equation}\label{equ:proof_2}
B^{\top } f_{\theta}(x) = B^{\top } f_{\tilde{\theta}}(x) + G
\end{equation}
where $G = \big(\log \tfrac{Z_{\tilde{\theta},j}}{Z_{\theta,j}}:j\in [k]\big)$ is a $k$-dimensional vector. By definition, $B B^{\top} = I_{k\times k}$. Then multiplying the two side of \eqref{equ:proof_2} by $B$ proves \cref{equ:id},
\begin{equation}\label{equ:claim}
f_{\theta}(x) = f_{\tilde{\theta}}(x) + C  \text{ for any } x\in \mathcal{X} \text{ and }  C= BG.
\end{equation}
Reversely, multiplying two sides of \eqref{equ:claim} by $\beta_j^{\top }$, we obtain \cref{equ:proof},
\[
\beta_j^{\top} f_{\theta}(x) = \beta_j^{\top} f_{\tilde{\theta}}(x) + \beta_{j}^{\top}B G =  \beta_j^{\top} f_{\tilde{\theta}}(x) + G_j
= \beta_j^{\top} f_{\tilde{\theta}}(x) + \log \tfrac{Z_{ \tilde{\theta},j}}{Z_{\theta,j}}
\]
Then multiplying $-1$ and applying $\exp(\cdot)$ to both sides,  
\[
 p_{\theta,j}(x) =Z_{\theta,j}^{-1}\exp\left[-\beta_j^{\top}  f_{\theta}(x)\right] =  Z_{\tilde{\theta},j}^{-1}\exp\left[-\beta_j^{\top}  f_{\tilde{\theta}}(x)\right] =  p_{\tilde{\theta},j}(x). 
\]
\end{proof}
The orthogonality of $B$ is also the key to prove the universal approximation capability of our partially randomized EBM in \Cref{prop:approx}; see the next section for more details.

\subsection{Proof of \Cref{prop:approx}}\label{sect:approx}

\begin{theorem}[Stone-Weierstrass; Theorem 4.45 in \citep{folland1999real}]\label{thm:stone} Suppose $\mathcal{X}=[0,1]^{d}$, and that $\mathcal{P}$ is a class of functions satisfying the following conditions:
\begin{enumerate}
    \item Every $p_X\in \mathcal{P}$ is continuous.
    \item For every $x\in \mathcal{X}$, there exists $p_X\in \mathcal{P}$ such that $p_X(x)\neq 0$.
    \item For every $x,x'\in \mathcal{X}$ such that $x\neq x'$, there exists $p_X\in \mathcal{P}$ such that $p_X(x) \neq p_X(x')$.
    \item $\mathcal{F}$ is closed under multiplication and under vector space operations.
\end{enumerate}
Then for every continuous density function $g:\mathcal{X}\rightarrow \mathbb{R}^+$ and $\epsilon>0,$ there exists $p_X\in \mathcal{P}$ s.t.
$\sup_{x\in\mathcal{X}}|g(x)-p_X(x)|\leq \epsilon.$
\end{theorem}
The original Stone-Weierstrass theorem works for any $g:\mathcal{X}\rightarrow \mathbb{R}$. Here we modify the codomain of $g$ to be $\mathbb{R}^{+}$ because we only consider estimating density functions. The Stone-Weierstrass theorem works for any $\mathcal X\subseteq \mathbb{R}^{d}$. Choosing $\mathcal X=[0,1]^{d}$ is common practice in the literature of approximation theory and nonparametric regression. It is used to simplify the notation in the proof. The proof idea holds for general $\mathcal X\subseteq \mathbb{R}^{d}$.

\begin{proof}

Suppose $f_{\theta}$ is a linear neural network in the model $\eqref{equ:model}$, we have
\[
p_{\theta,j} \in \mathcal{P}(\beta_j):= \left\{x\rightarrow Z_{\theta,j}^{-1}\exp\left[-\beta_j^{\top}  \theta x\right]:\theta \in \mathbb{R}^{k\times d}\right\}
\]
The first condition in \Cref{thm:stone} is satisfied directly by definition. If $\theta$ is a zero matrix, then $p_{\theta,j}(x) \neq 0$, which means the second condition in \Cref{thm:stone} holds. Given any $x_1,x_2\in \mathcal{X}$ such that $x_1\neq x_2$, we can always find a exponential function along the line between $x_1$ and $x_2$. Because the exponential function  $p_{\theta,j} \in \mathcal P(\beta_j)$ is nonlinear, we have $p_{\theta,j}(x_1)\neq p_{\theta,j}(x_2)$. Formally, we let $\theta = -\left(\sum_{j'=1}^{k}\beta_{j'}\right) (x_1-x_2)^{\top }$, then 
\[
-\beta_j^{\top}\theta =1\cdot(x_1-x_2)^{\top } = (x_1-x_2)^{\top },
\]
The first equality uses the fact that $\beta_j$ is a column of the random orthogonal matrix $B$. Then,
\[
p_{\theta,j}(x) = Z_{\theta,j}^{-1}\exp\left[(x_1-x_2)^{\top }x\right].
\]
Because $x_1\neq x_2$, we have
\[
\frac{p_{\theta,j}(x_1)}{p_{\theta,j}(x_2)} = \frac{\exp\left[x_1^{\top }x_1-x_2^{\top }x_1\right]}{\exp\left[x_1^{\top }x_2-x_2^{\top }x_2\right]} = \exp\left[ -(x_1-x_2)^{\top }(x_1-x_2)\right]>0,
\]
which proves that that the third condition in \Cref{thm:stone} holds. The space $\mathcal{P}(\beta_j)$ is closed under vector space operations by definition. It is also closed under multiplication. By the property of exponential functions, $\exp\left[-\beta_j^{\top}  \theta x_1\right]\cdot \exp\left[-\beta_j^{\top}  \theta x_2\right] = \exp\left[-\beta_j^{\top}  \theta (x_1+x_2)\right]$, it is straightforward to show that given some functions $p$'s $\in \mathcal{P}(\beta_j)$, the multiplication of two different linear combinations of $p$'s is still in the span of $p$'s. Thus, $\mathcal{P}(\beta_j) $ satisfies the fourth condition in \Cref{thm:stone}.
\end{proof}

Although the derivation is based on a linear network $f_{\theta}$ for simplicity, it does not mean we should use a linear network in practice. In recent years, it has been shown theoretically and empirically that overparametrization in deep neural networks allows gradient methods to find interpolating solutions; these methods implicitly impose regularization; overparametrization leads to benign overfitting, that is, accurate predictions (i.e. better generalization) on the testing data despite overfitting training data; see \citep{bartlett2021deep} for a detailed review of the recent theoretical analysis on overparameterized models.

\subsection{Proof of \Cref{prop:consistent}}\label{sect:consistent}

\begin{proof}
As discussed in the main paper, both MLE and NCE are M-estimators.  The proof consists of two steps: (1) we show $q_{\theta,j}(a \mid V_i)$ with $p_{\theta,j}(x)= p_X(x)$
is a maximizer of $ \mathcal{L}_{\infty,j}(\theta)$, then (2)
we show the standard conditions of consistent M-estimators hold for $\mathcal{L}_{n,j}(\theta) $. 
We assume the sample splitting always keep the individuals in the same folds as $n\rightarrow\infty$. This can be achieved by keeping the existing individuals in the same folds and randomly assign a new individual to a fold as $n \rightarrow \infty$. For large $n$, each fold will have roughly the same number of individuals, and $n_j= |\mathcal{I}_j|\rightarrow \infty $ for every $j\in [k]$ as $n\rightarrow\infty$.
The assumption does not affect our method in practice because we only consider the number of individuals $n$ observed in our dataset.


\textbf{Step 1.} Recall that for every $i\in [n]$, 
$\bar{X}_i = (X_i,\tilde{X}_{i1},\dotsc, \tilde{X}_{ib} )\sim  p_X(x)\prod_{a=1}^{b}\tilde{p}_{\tilde{X}\mid X}(\tilde{x}\mid x)$. We randomly permute the columns of $\bar{X}_i$ and let $V_i = (V_{i,1},\cdots V_{i,b+1})$ be the permuted $\bar{X}_i$. Each column of $V_i$ has equal probability $(b+1)^{-1}$ for being the clean sample $X_i$. The variable $W_i\in \{0,1\}^{b+1}$ indicate which column of $V_i$ is $X_i$. Here, we define a categorical variable $S_i\in [b+1]$ such that $S_i = a$ if $W_{ia} =1 $. We know that
\[
p_{S_i}(a )  = 1/(b+1),~\forall a\in [b+1].
\]
We define the marginal distribution of $v=(v_c:c\in[b+1])$ as
\begin{align*}
\Lambda_j (v) = \sum_{a=1}^{b+1}p_S(a ) p_{V\mid S}(v|a)=  \sum_{a=1}^{b+1} (b+1)^{-1} p_{X}(v_{a})\tilde{p}_{-a}(v) 
\end{align*}
where  $\tilde{p}_{-a}(v) =\prod_{a\in [b+1]:a'\neq a}p_{\tilde{X}\mid X}(v_{a'}\mid v_{a})$.
We define the posterior probability of $S=a$  as
\[
p_{S\mid V}(a\mid v) = \frac{(b+1)^{-1} p_{X}(v_{a}) \tilde{p}_{-a}(v)         }{\sum_{c=1}^{b+1} (b+1)^{-1} p_{X}(v_{c})\tilde{p}_{-c}(v)  }
\]
This corresponds to the posterior probability $q_{\theta,j}(a \mid v) $ based on the model $p_{\theta,j}(x)$  in \eqref{equ:poster_prob}.

As $n\rightarrow\infty$, i.e., $n_j\rightarrow\infty$, the objective function in \eqref{equ:rank} is given by 
\begin{align*}
\mathcal{L}_{\infty,j}(\theta) 
 = \int \Lambda_j (v) \left[ \sum_{a=1}^{b+1}   p_{S\mid V}(a\mid v) \log q_{\theta,j}(a \mid v)  \right]  d v \\
\end{align*}
Because $\Lambda_j(v) >  0 $ and \Cref{lemma:B_2},
$\mathcal{L}_{\infty,j}(\theta) $ is maximized when 
\begin{equation}\label{equ:expand}
q_{\theta,j}(a \mid v)= p_{S\mid V}(a\mid v),~\forall a\in [b+1].
\end{equation}
Suppose  $v =(v_{a'}:a' \in [b+1])$ satisfies that $  v_{a'}= \xi$ for all $a'\in [b+1]\setminus\{a\}$. Then
\[
\tilde{p}_{-c}(v) = \tilde{p}_{\tilde{X}\mid X}(v_a \mid \xi )  \left[p_{\tilde{X}\mid X}(\xi \mid \xi )\right]^{b-1}, \forall c\in [b+1]\setminus\{a\}
\]
and 
\begin{equation}\label{equ:c_c'}
\tilde{p}_{-c}(v) = \tilde{p}_{-c'}(v), \forall c,c'\in [b+1]\setminus\{a\}.
\end{equation}
We continue to rewrite \eqref{equ:expand} as
\begin{align*}
 \frac{p_{\theta,j}(v_{a}) \tilde{p}_{-a}(v)}{\sum_{c=1}^{b+1} p_{\theta,j}(v_{c}) \tilde{p}_{-c}(v) }  &=  \frac{p_{X}(v_{a})\tilde{p}_{-a}(v)}{\sum_{c'=1}^{b+1} p_{X}(v_{c'}) \tilde{p}_{-c'}(v)}    \\
 \frac{p_{\theta,j}(v_{a})}{\sum_{c=1}^{b+1} p_{\theta,j}(v_{c}) \tilde{p}_{-c}(v)}  &=  \frac{p_{X}(v_{a})}{\sum_{c'=1}^{b+1} p_{X}(v_{c'}) \tilde{p}_{-c'}(v)}    \\
 \frac{ \sum_{c=1}^{b+1} p_{\theta,j}(v_{c}) \tilde{p}_{-c}(v) }{p_{\theta,j}(v_{a})  }  &=  \frac{\sum_{c'=1}^{b+1} p_{X}(v_{c'}) \tilde{p}_{-c'}(v)}{p_{X}(v_{a})}    \\
 \tilde{p}_{-a}(v_{a})   + \sum_{c\neq a } \frac{  p_{\theta,j}(v_{c}) \tilde{p}_{-a}(v_{c})  }{  p_{\theta,j}(v_{a}) } & =  \tilde{p}_{-a}(v_{a})  +\sum_{c'\neq a }  \frac{ p_{X}(v_{c'}) \tilde{p}_{-a}(v_{c'}) }{  p_{X}(v_{a})   }   \\
 \frac{  p_{\theta,j}(\xi) }{  p_{\theta,j}(v_{a}) } & =    \frac{ p_X(\xi) }{ p_{X}(v_{a})   }    \\
\beta_{j}^{\top}\left[f_{\theta}(\xi) - f_{\theta}(v_a)\right] & = \beta_{j}^{\top}\left[f_{\theta_0}(\xi) - f_{\theta_0}(v_a)\right],
\end{align*}
The fifth line is attained by \eqref{equ:c_c'}. The last line is achieved by the setup of our proposition: $p_{\theta_0,j}(x) = p_X(x)$ for any  $\theta_0 \in \Theta_0.$ 
Combing the last equation for all $j\in [k]$ and using the fact that $B$ is a orthogonal matrix, we have
\begin{align*}
B^{\top}\left[f_{\theta}(\xi) - f_{\theta}(v_a)\right] &=
B^{\top}\left[f_{\theta_0}(\xi) - f_{\theta_0}(v_a)\right]\\
f_{\theta}(\xi) - f_{\theta}(v_a) &  = f_{\theta_0}(\xi) - f_{\theta_0}(v_a)   \\
  f_{\theta}(v_a)   &  = f_{\theta_0}(v_a) +  C(\theta,\theta_0)
\end{align*}
Then,
\begin{align*}
  p_{\theta,j}(x) =Z_{\theta,j}^{-1}\exp\left[-\beta_j^{\top}  f_{\theta}(x)\right] 
  & = \frac{\exp\left[-\beta_j^{\top}  f_{\theta_0}(x) -\beta_j^{\top} C(\theta,\theta_0) \right]}{\int_{\mathcal{X}} \exp\left[-\beta_j^{\top}  f_{\theta_0}(x) -\beta_j^{\top} C(\theta,\theta_0) \right] dx}  =   p_X(x). \\
\end{align*}
For any $\hat{\theta}\in \argmax_{\theta\in \Theta} \mathcal{L}_{\infty,j}(\theta)$, we have $p_{\hat{\theta},j}(x) = p_X(x)$ for any $x\in \mathcal{X}$. 

\textbf{Step 2.} Because $\mathcal{X}$ and $\Theta$ are compact, the value of each covariate and network parameter is bounded. Then the function we optimize in \cref{equ:rank} is
\[
g_j(\bar{x};\theta) =\log q_{\theta,j}(1 \mid \bar{x})
\]
is bounded for any $\bar{x} = (x,\tilde{x}_{i1},\dotsc,\tilde{x}_{ib})$.  Then we denote $\mathcal{L}_{n,j}(\theta)$ in \cref{equ:rank} by $\mathbb{E}_{n} \left[ g_j(\bar{X};\theta)  \right]  = n_j^{-1}\sum_{i\in \mathcal{I}_j}
g_j(\bar{X}_i;\theta) $. Using the uniform law of large number (ULLN) (see \citep[Theorem 2]{jennrich1969asymptotic} and  \citep[Lemma 2.4]{newey1994large}), we have
\[
\sup_{\theta\in \Theta} \left|  \mathbb{E}_{n} \left[ g_j(\bar{X};\theta)  \right]  - \mathbb{E}\left[ g_j(\bar{X};\theta)  \right]\right|\overset{p}{\to} 0.
\] 
Now we change the exact identifiability assumption in
\citep[Theorem 12.2]{wooldridge2010econometric} 
and \citep[Theorem 2.5]{newey1994large} to our partial identifiability assumption. Under \Cref{prop:approx}, we suppose that there is a countable subset $\Theta_0\subset \Theta$ such that for every $\theta_0\in \Theta_0$, $p_{\theta_0,j}(x)$  gives the same distribution as $p_{X}(x)$, and any $\theta_0$ is a non-unique maximizer of $\mathbb{E}\left[ g_j(\bar{X};\theta)\right]$. We define an open ball with radius equal to $\eta>0$ for every $\theta_0 \in \Theta_0$. The region inside and outside these open balls is given by
\[
\Theta_{\eta} = \left\{\theta\in \Theta  \ \bigg|  \  \argmin_{\theta_0\in \Theta_0} \|\theta-\theta_0 \|_2 < \eta \right\}
\quad \text{and} \quad 
\Theta_{\eta}^{\text{c}} = \left\{\theta\in \Theta  \ \bigg|  \  \argmin_{\theta_0\in \Theta_0} \|\theta-\theta_0 \|_2 \geq \eta \right\}.
\]
Using the proof of
\citep[Theorem 2.1]{newey1994large}, $\forall \epsilon>0$, $\theta_0\in \Theta_0$ and $\hat{\theta}_n \in \argmax \mathbb{E}_{n} \left[ g_j(\bar{X};\theta)  \right]$, we have with probability approaching to 1:
\begin{equation}\label{equ:bound}
\mathbb{E} \left[ g_j(\bar{X};\hat{\theta}_n)  \right] > \mathbb{E} \left[ g_j(\bar{X};\theta_0) \right]   -\epsilon. 
\end{equation}
By the compactness of $\Theta_{\eta}^{\text{c}}$ and the assumption that $f_{\theta}$ is continuous w.r.t to $\theta$, we have
\[
\sup_{\theta\in \Theta_{\eta}^{\text{c}}}\mathbb{E} \left[ g_j(\bar{X};\theta)  \right] = \mathbb{E} \left[ g_j(\bar{X};\theta^*)  \right]  < \mathbb{E} \left[ g_j(\bar{X};\theta_0)  \right] \text{ for some } \theta^* \in \Theta_{\eta}^{\text{c}}.
\]
Thus, by
$\epsilon = \mathbb{E} \left[ g_j(\bar{X};\theta_0)  \right] - \sup_{\theta\in \Theta_{\eta}^{\text{c}}}\mathbb{E} \left[ g_j(\bar{X};\theta)  \right],$ it follows from \eqref{equ:bound} that with probability approaching to 1,
\begin{equation}\label{equ:end}
\mathbb{E} \left[ g_j(\bar{X};\hat{\theta}_n)  \right] >  \sup_{\theta\in \Theta_{\eta}^{\text{c}}}\mathbb{E} \left[ g_j(\bar{X};\theta)  \right] \Rightarrow 
\hat{\theta}_n \in \Theta_{\eta}.
\end{equation}
Since \eqref{equ:end} is true for any $\eta>0$, we have $\hat{\theta}_n\in \Theta_0$ with probability 1 as $n\rightarrow \infty$. We note that the same proof holds if we consider the summation of $\mathbb{E}_{n} \left[ g_j(\bar{X};\theta)  \right]$ over all $j\in [k]$. Therefore, for any number of noise samples $b$ and
$\hat{\theta}_n \in \argmax_{\theta\in\Theta} \mathcal{L}_{n}(\theta) $,  $\lim_{n\rightarrow\infty}\hat{\theta}_n \in   \Theta_0 $ with probability 1. 
\end{proof}

\subsection{Supporting Lemma}

\begin{lemma}\label{lemma:B_2}
Suppose $w = (w_1,\dotsc,w_b)> 0 $ and $\sum_{a=1}^{b}w_a=1$, 
\[
f(\tilde{w};w) = \sum_{a=1}^{b}w_a\log \tilde{w}_a \quad \text{subject to}\quad  \tilde{w} = (\tilde{w} _1,\dotsc,\tilde{w} _b)> 0 \text{ and } \sum_{a=1}^{b}\tilde{w}_a=1.
\]
\end{lemma}
Then $f(\tilde{w};w)$ is maximized at $\tilde{w} = w$.
\begin{proof}
Suppose 
$
g(\tilde{w};w)  = \sum_{a=1}^{b}w_a\log \tilde{w}_a  + \lambda \left(\sum_{a=1}^{b}\tilde{w}_a - 1 \right).
$ We have
\[
\frac{\partial g(\tilde{w};w)}{\partial \tilde{w}_c} = 0 \Rightarrow  \tilde{w}_c= - \frac{w_c}{\lambda}  \quad \text{and} \quad 
\frac{\partial g(\tilde{w};w)}{\partial \lambda} = 0 \Rightarrow  \sum_{a=1}^{b}\tilde{w}_a = 1.
\]
Combining both conditions, we have
$\sum_{a=1}^{b}\tilde{w}_a = - \frac{1}{\lambda}\sum_{a=1}^{b}w_c = - \frac{1}{\lambda} = 1 \Rightarrow \lambda = -1$.
Then, $ \tilde{w}_c = - \frac{w_c}{ -1} \Rightarrow \tilde{w}_c = w_c$. Then by a second-derivative test on the bordered Hessian of $g(\tilde{w};w)$, we have $\tilde{w} =  (\tilde{w}_1,\dotsc,\tilde{w}_b) = (w_1,\dotsc,w_b) $ is a maximizer of the function $f(\tilde{w};w)$.
\end{proof}

\subsection{The noise distribution $\tilde{p}(\tilde{x} \mid X_i)$}\label{sect:noise}
We draw a noise sample from $p_{\tilde{X}\mid X}(\tilde{x}\mid X_i)$.
For each feature $X_{is}$ of the $d$-dimensional $X_i$, $s\in[d]$, we sample an independent binary variable $R_{is}\sim \text{Ber}(q)$ to decide if the $s$-th feature $X_{is}$  will be corrupted. If $R_{is}=1$, we will corrupt  the $s$-th feature, otherwise not. Overall, the first part of $p_{\tilde{X}\mid X}(\tilde{x}\mid X_{i})$ is given by
\[
\prod_{s=1}^{d}q^{R_{is}}(1-q)^{1-R_{is}},
\]
where $q$ is the only hyperparameter in $p_{\tilde{X}\mid X}(\tilde{x}\mid X_{i})$. 
We use the same $q$ for all $j\in [k]$. In \Cref{sect:app:hyper_details}, we describe how
$q$ (called perturbation prob.) is selected by validation in our experiments.

Suppose that  $R_{is}=1$. If the $s$-th feature is continuous,  we will corrupt it by adding a white noise $E_{is}$ drawn from a standard normal distribution, i.e.,
\[
E_{is}\sim p_{E_s\mid R_{s}}(e_s\mid 1)  = \mathcal{N}(0,1).
\]
If the $s$-th feature is categorical and takes its value in $\mathcal{X}_s$, we will corrupt it by replacing $X_{is}$ with a uniform sample $E_{is}$  drawn from the same range $\mathcal{X}_s$, i.e.,
\[
E_{is}\sim p_{E_s\mid R_{s}}(e_s\mid 1) =  1/|\mathcal{X}_s|.
\]
Suppose that $R_{is}=0$. We do not corrupt the $s$-th feature. That is,
\begin{itemize}
    \item $E_{is}=0 $ and $p_{E_s\mid R_{s}}(0\mid  0)=1$ if the $s$-th feature is continuous and $R_{is}=0$;
       \item $E_{is}=X_{is}$ and $p_{E_s\mid R_{s}, X_s}(X_{is}\mid 0,X_{is}) = 1$ if the $s$-th feature is categorical and $R_{is}=0$.
\end{itemize}
Overall, the probability $p_{\tilde{X}	\mid X}(\tilde{X}_i\mid X_i)$ is computed using  $R_{is}$, $E_{is}$ and $X_{is}$ for all $s\in [d]$,
\[
p_{\tilde{X}\mid X}(\tilde{X}_i\mid X_i) = \prod_{s=1}^{d}q^{R_{is}}(1-q)^{1-R_{is}}p_{E_s	\mid R_{s},X_s}(E_{is}\mid R_{is},X_{is}).
\]
where $p_{E_s\mid R_{s},X_{s}}(E_{is}\mid R_{is},X_{is})$ only depends on $X_{is}$ if the $s$-th feature is categorical and $R_{is}=0$, otherwise $p_{E_s	\mid R_{s},X_s}(E_{is}\mid R_{is},X_{is}) = p_{E_{s}\mid R_{s}}(E_{is}\mid R_{is})$.
The corrupted sample $\tilde{X}_{i}$ is obtained by either adding $E_{is}$ to $X_{is}$ or replacing $X_{is}$ by $E_{is}$ for every $s\in [d].$  We do not need to consider this step when we compute $\tilde{p}_{\tilde{X}\mid X}(\tilde{X}_{i}\mid X_{i})$.

\end{document}